# FAST COMPUTATION OF PERCLOS AND SACCADIC RATIO

*Anirban Dasgupta*

# Fast computation of PERCLOS and Saccadic Ratio

**A dissertation submitted in partial fulfilment of the requirements**
*to the Indian Institute of Technology, Kharagpur*
*For award of the degree*

*of*

**Master of Science (by Research)**

*by*

**Anirban Dasgupta**

**Roll No.: 10EE70P07**

**Under the esteemed guidance of**

**Professor Aurobinda Routray**

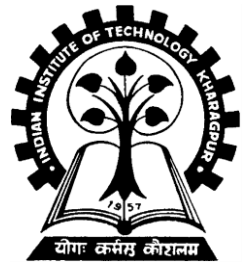

**DEPARTMENT OF ELECTRICAL ENGINEERING**

**INDIAN INSTITUTE OF TECHNOLOGY, KHARAGPUR**

**SEPTEMBER 2013**

# <u>CERTIFICATE</u>

This is to certify that the thesis entitled **Fast computation of PERCLOS and Saccadic Ratio** submitted by **Anirban Dasgupta** to Indian Institute of Technology Kharagpur, is a record of bona fide research work under our supervision and I consider it worthy of consideration for the award of the degree of Master of Science by Research of the Institute.

_______________________________

Supervisor

Prof. Aurobinda Routray

Department of Electrical Engineering

Indian Institute of Technology, Kharagpur

Kharagpur, West Bengal

India - 721302

Date:

# ACKNOWLEDGEMENTS

I wish to express my deep sense of gratitude and indebtedness to my supervisor Prof. Aurobinda Routray, Professor, Department of Electrical Engineering, Indian Institute of Technology, Kharagpur for his constant supervision, guidance and invaluable suggestions during the research work. I express my sincere thanks to Prof. S. Sen, Head of the Department, Prof. J. Pal, Ex. Head of the Department, Prof. A. K. Deb, Prof. A. Mukherjee and Prof. P. K. Dutta, Department of Electrical Engineering for their suggestions which have helped me improve the quality of my work. It is also my pleasure to express my deep sense of gratitude and indebtedness to Head of the Department, Department of Electrical Engineering, Indian Institute of Technology, Kharagpur for providing facilities and resources in the department for pursuing my research work.

I gratefully acknowledge the funds received from Department of Electronics and Information Technology, Ministry of Information Technology, Government of India and Samsung Global Research Outreach 2012.

The suggestions obtained from my project review members – Prof. A. Konar, Prof. S. Dandapat, Prof. A. Rakshit, Prof. N. Srinivasan, Captain A.P. Kannan and Mr. Tara Shanker were really valuable.

I would also like to take this opportunity to express my appreciation for Dr. Supratim Gupta, Dr. Shibsambhu Kar, Mr. S L Happy, Mr. Anjith George, Mr. Aritra Chaudhuri, Miss Anwesha Sengupta, Mr. Tapan Pradhan, Mr. Bibek Kabi, Miss Aparna Rajaguru, Mr. Prasenjit Das, Mr. M. Sai Nataraj, Mr. Manish Singhvi, Mr. R. Vignesh, Mr. Ganesh Bhalsingh, Mr. Anik Kumar Samanta, Mr. Arunava Naha, Miss Soumi Chaki, Miss Anushree Basu, Mr. Ayan Mukherjee, Mr. Manas Midya, Mr. Shubhasis Bhowmick, Mr. Aniruddha Das, Mr. Binay Rava and all other friends who have encouraged me. I wish all of you a lot of success in your professional and personal endeavours. I would like to thank the subjects who voluntarily participated in the experiments for data collection.

This work would not have been accomplished without the support and love of my family members. I would remain totally indebted to them for their encouragement.

Place: IIT Kharagpur

Date:                                                          Anirban Dasgupta

# DECLARATION

I certify that

a. The work contained in the thesis is original and has been done by myself under the general supervision of my supervisor(s).

b. The work has not been submitted to any other Institute for any degree or diploma.

c. I have followed the guidelines provided by the Institute in writing the thesis.

d. I have conformed to the norms and guidelines given in the Ethical Code of Conduct of the Institute.

e. Whenever I have used materials (data, theoretical analysis, and text) from other sources, I have given due credit to them by citing them in the text of the thesis and giving their details in the references.

f. Whenever I have quoted written materials from other sources, I have put them under quotation marks and given due credit to the sources by citing them and giving required details in the references.

Signature of the Student

# Abstract

This thesis describes the development of fast algorithms for computation of PERcentage CLOSure of eyes (PERCLOS) and Saccadic Ratio (SR). PERCLOS and SR are two ocular parameters reported to be measures of alertness levels in human beings. PERCLOS is the percentage of time in which at least 80% of the eyelid remains closed over the pupil. Saccades are fast and simultaneous movement of both the eyes in the same direction. SR is the ratio of peak saccadic velocity to the saccadic duration. This thesis addresses the issues of image based estimation of PERCLOS and SR, prevailing in the literature such as illumination variation, poor illumination conditions, head rotations etc. In this work, algorithms for real-time PERCLOS computation has been developed and implemented on an embedded platform. The platform has been used as a case study for assessment of loss of attention in automotive drivers. The SR estimation has been carried out offline as real-time implementation requires high frame rates of processing which is difficult to achieve due to hardware limitations. The accuracy in estimation of the loss of attention using PERCLOS and SR has been validated using brain signals, which are reported to be an authentic cue for estimating the state of alertness in human beings. The major contributions of this thesis include database creation, design and implementation of fast algorithms for estimating PERCLOS and SR on embedded computing platforms.

| Symbol | Description |
|--------|-------------|
| $P$ | PERCLOS |
| $E_c$ | closed eye count over a predefined interval |
| $E_t$ | total eye count in the same given interval |
| $SR$ | Saccadic Ratio |
| $V_p$ | Peak Saccadic Velocity |
| $T_s$ | Saccadic Duration |
| $\gamma(x, y)$ | Correlation co-efficient |
| $i(x, y)$ | Image Intensity |
| $I(x, y)$ | Integral Image |
| $h_j(x)$ | Weak classifier based on Haar-like features |
| $K$ | No. of classifiers |
| $d_i$ | Detection rate of the $i^{th}$ Haar classifier |
| $D$ | Overall detection rate of the Haar classifier |
| $f_i$ | false positive rate of the $i^{th}$ Haar classifier |
| $F$ | Overall false positive rate of the Haar classifier |
| tp | No. of true positives |
| fp | No. of false positives |
| fn | No. of false negatives |
| tn | No. of true negatives |
| $SF$ | Scale Factor |
| $R$ | Rotation Matrix |
| $\theta$ | Angle of rotation of affine rotation matrix |
| $LBP$ | Local Binary Pattern Value |
| $IPF_v$ | Vertical Integral Projection Function |
| $IPF_h$ | Horizontal Integral Projection Function |
| $VPF_v$ | Vertical Variance Projection Function |
| $VPF_h$ | Horizontal Variance Projection Function |
| $GPF_v$ | Vertical Generalized Projection Function |
| $GPF_h$ | Horizontal Generalized Projection Function |
| $\alpha$ | Parameter for Hybrid Projection Function |
| $z_k$ | Measurement at $k^{th}$ instant |
| $H$ | Output transition matrix |
| $x_k$ | State vector at $k^{th}$ instant |
| $v_k$ | Measurement noise at $k^{th}$ instant |
| $A$ | State transition matrix |
| $B$ | Input transition matrix |
| $u_k$ | Control input at $k^{th}$ instant |
| $w_k$ | Process noise at $k^{th}$ instant |
| $Q$ | Covariance matrix for Process noise |
| $R$ | Covariance matrix for Measurement noise |

| Abbreviation | Description |
| --- | --- |
| PERCLOS | PERcentage of CLOSure |
| AECS | Average Eye-Closure Speed |
| SR | Saccadic Ratio |
| EEG | Electro Encephalo Gram |
| EOG | Electro Oculo Gram |
| AdaBoost | Adaptive Boosting |
| SF | Scale Factor |
| ROI | Region Of Interest |
| ROC | Receiver Operating Characteristics |
| PCA | Principal Component Analysis |
| LBP | Local Binary Patterns |
| SVM | Support Vector Machines |
| KF | Kalman Filter |
| CKF | Constrained Kalman Filter |
| LDA | Linear Discriminant Analysis |
| BHE | Bi-Histogram Equalization |
| CRP | Corneal Retinal Potential |
| AHE | Adaptive Histogram Equalization |
| CLAHE | Contrast Limited Adaptive Histogram Equalization |
| DoG | Difference of Gaussians |
| IPF | Integral Projection Function |
| VPF | Variance Projection Function |
| GPF | Generalized Projection Function |
| HPF | Hybrid Projection Function |
| BPDP | Bright Pupil - Dark Pupil |
| NIR | Near Infra-Red |
| fps | frames per second |
| ppf | pixels per frame |
| SBC | Single Board Computer |
| MA | Maximum Angle |
| MD | Maximum Deviation |
| LDR | Light Dependent Resistor |
| PSD | Power Spectral Density |
| FF | Form Factor |
| RMS | Root Mean Square |
| ESI | Edge Strength Index |
| CHT | Circular Hough Transform |
| AWGN | Additive White Gaussian Noise |
| GPU | Graphics Processing Unit |

# Contents

# List of Figures

## List of Tables

# Chapter 1. Introduction – PERCLOS and Saccade

## 1.1 The Motivation

There have been some studies [1] [2] [3] with regard to ocular parameters such as PERcentage of eyelid CLOSure (PERCLOS) and Saccadic Ratio (SR) as indicators of alertness in human beings. Image based estimation of these parameters has been a challenge to the research community over the years.

## 1.2 Introduction

Alertness level in human beings can be assessed using different measures such as Electro Encephalo Gram (EEG) [4, 5], blood samples [6], ocular features [2], speech signals [7] and skin conductance [5]. The EEG signals and blood samples based methods have been reported to be most accurate for estimating the state of drowsiness [4, 8]. These methods being contact-based limits their feasibility of implementation on practical scenarios. There have been some studies [9], [10], [11], [12], [13] with regard to eyelid movements such as blink frequency, Average Eye-Closure Speed (AECS), PERCLOS, eye saccade etc. as quantitative measures of the alertness level of an individual. Of them, PERCLOS is reported to be the best and most robust measure for drowsiness detection [14], [15] whereas saccadic movements are reported to be the best indicator of the alertness level [16]. PERCLOS is based on eye closure rates whereas SR is based on fast eye movements. The computation of PERCLOS and Saccadic Ratio (SR) on real-time platforms may be useful for monitoring alertness levels of individuals who have to be alert during their jobs, such as air traffic controllers, pilots or operators in power-grids, long-distance automotive and locomotive driving, medical health care etc. This thesis presents a case study for automotive driving only. However, the system can be used for in all the above mentioned cases with suitable implementation.

## 1.3 PERCLOS

PERCLOS as a drowsiness metric was established by Wierwille *et al.* [10]. PERCLOS [17] may be defined as the proportion of time the eyelids are at least 80% closed over the pupil. PERCLOS can be estimated from continuous video sequences of the eye images [18]. There are several steps involved before estimating the accurate PERCLOS value. The first step is face detection followed by eye detection and eye state classification. Finally, the PERCLOS value is calculated as follows.

$$P = \frac{E_c}{E_t} \qquad\qquad 1\text{-}1$$

In the above equation, $P$ is the PERCLOS value, $E_c$ is the number of closed eye count over a predefined interval while $E_t$ is the total eye count in the same given interval. For e.g. if 100 frames of facial images are recorded and 20 such cases are found where the eyes are closed (at least 80% spatially), then the PERCLOS value is 20%. The eyes are categorised as open or closed during training based on 80% spatial closure. The details are provided in Section 3.6. Literature [1] states that a higher value of $P$ indicates higher drowsiness level and vice versa. From 1-1, it is apparent that to compute $P$, it is necessary to detect and classify the eye states accurately.

Several researchers have reported about PERCLOS as an indicator of drowsiness level. Hong *et al.* [19] developed an embedded system to detect driver's drowsiness by computing the PERCLOS. They detected face using Haar-like features and localized the eye using horizontal and vertical variance projection. They used Camshift algorithm to track the eyes, thereby limiting the search area by Camshift prediction in YCbCr plane. They classified the eye state by thresholding eye region, then applying Laplacian on the binary image and finally using a complexity function. They obtained an accuracy of about 85%. The algorithm has been tested for real-time applications. Being dependent on color information, the algorithm is illumination dependent. Its performance may be poor for on-board situations where occurrence of shadows will cause problems. Robotics Institute in Carnegie Mellon University has developed a drowsy driver monitor named Copilot [20], which is a video-based system and estimates PERCLOS. However the issues such as illumination variation and head rotations have not been adequately addressed. Hayami *et al.* [21] compared PERCLOS with vertical eye movement frequencies between wakeful and drowsy states of drivers in a driving simulator for three drivers. Bergasa *et al.* [22] have used ellipse fitting for detecting the eye state and computing PERCLOS. They have used active illumination methods for locating the eyes. In [23], Ji *et al.* describe a real-time system for monitoring driver's attention level by tracking the eyes and computing PERCLOS. They have employed active illumination techniques for tracking the eyes. A major drawback of active illumination techniques is its harmful effect on sustained exposure on the eyes. In this work, passive illumination has been used to estimate PERCLOS.

## 1.4 Eye Detection – A Review

From the above review, it is apparent that for the accurate estimation of PERCLOS, fast and accurate detection of eyes is necessary. Several approaches on eye detection have been reported in literature. Orazio *et al.* [24] applied a neural network based classifier to detect the eyes. The classifier contained 291 eye images and 452 non-eye images as training. The algorithms are tested offline. In [25], Huang *et al.* used optimal wavelet packets for representing eyes, thereby classifying the face area into eye and non-eye regions using Radial Basis Functions. The optimal wavelet packets are selected by minimizing their entropy. They have obtained an eye detection rate of 82.5% for eyes and 80% for non-eyes using 16 wavelet coefficients whereas 85% for eyes and 100% for non-eyes using 64 wavelet coefficients. The real-time performance of the algorithm has not been established in the work. In [26], Feng *et al.* employed three cues for eye detection using gray intensity images. These are face intensity, the estimated direction of the line joining the centers of the eyes and the response of convolving the proposed eye variance with the face image. Based on the cues, a cross-validation process has been carried out to generate a list of possible eye window pairs. For each possible case, the variance projection function is used for eye detection and verification. They obtained a detection accuracy of 92.5% on a face database from MIT AI laboratory, which contains 930 face images with different orientations and hairstyles captured from different persons. However, low accuracy in eye detection was obtained in situations such as rotation of the face or occlusion of eyes due to eyebrows. Besides, this work does not address the real time implementation issues. In [27], Asteriadis *et al.* used pixel to edge information for eye detection. The edge map of the face is first extracted and a vector is assigned to every pixel, which is pointing to the closest edge pixel. Length and slope information for these vectors is accordingly used to detect and localize the eyes. The algorithm has been tested on Bio-ID and XM2VTS database. Real-time implementation of the algorithm has not been explored. In [28], Sirohey *et al.* used two methods of eye detection in facial images. The first is a linear method using filters based on Gabor wavelets. They obtained a detection rate 80% on Aberdeen dataset and 95% on the dataset prepared by them. The second was a nonlinear filtering method to detect the corners of the eyes using color-based wedge shaped filters, which has a detection rate of 90% on the Aberdeen database. The nonlinear method performed better over the linear method in terms of false alarm rate. Real-time realizations of the methods have not been discussed. Zhou *et al.* [11] defined the Generalized Projection Function (GPF) as a weighted combination of the Integral Projection Function (IPF) and the Variance Projection Function (VPF). With $\alpha$ and (1- $\alpha$) as weights of VPH and IPF respectively, they proved that IPF and VPF are special cases of GPF when the parameter $\alpha$ is 0 and 1, respectively. They also got Hybrid Projection Function (HPF) from GPF using $\alpha$ as 0.6. From HPF, they found that eye area is darker than its neighboring areas and the intensity of the eye area rapidly changes. They achieved best detection rates with $\alpha = 0.6$, on BioID, JAFFE and NJU face databases with a detection rate of 94.81%, 97.18% and 95.82% respectively. In these cases, the detection rate used in the survey is the true positive rate (TPR) conditioned that the error between detected eye center and manually marked eye center as a fraction of the eye width is less than a threshold for a detection to be considered as a true positive. In [18], Song *et al.* presented a new eye detection method which consists of extraction of Binary Edge Images (BEIs) from the grayscale face image using multi-resolution wavelet transform. They achieved an eye detection rate of 98.7% on 150 Bern images with variations in views and gaze directions and 96.6% was achieved on 564 images in AR face databases with different facial expressions and lighting conditions. The algorithm has not been tested in real-time and is prone to illumination variation as it uses intensity information for eye localization. The illumination variation issues have not been properly addressed in this work. In [29], Wang *et al.* studied the impact of eye locations on face recognition accuracy thereby introducing an automatic technique for eye detection. They used Principal Component Analysis (PCA) and PCA combined with Linear Discriminant Analysis (LDA) technique for face recognition. Subsequently, AdaBoost was used to learn the discriminating features to detect eyes. They tested their technique on FRGC 1.0 database and achieved an overall accuracy of 94.5% in eye detection. They also found that only PCA recognizes faces better than PCA combined with LDA. However, their technique has been tested for offline applications. Liying *et al.* [30] detected the eyes, based on skin color segment and identified the eye's condition by combining the color segmentation and morphological operations. However, the algorithm being dependent on color information will also be prone to variations in lighting conditions. Face rotation is also not considered in the work. Hansen *et al.* [31] developed an improved likelihood for eye tracking in uncontrolled environments. They have used BPDP method to detect the eyes. Li *et al.* [32] propose a fuzzy template matching based eye detection. Five feature parameters have been defined as a measure of similarity and calculated based on the template. The method is highly illumination dependent and variant with distance and orientation of the subject from the camera. In [9], Smith *et al.* used global motion and colour statistics to track a person's head and facial features to detect eyes thereby obtaining 74.8% detection rate on an average basis on different sets. Nevertheless, there are some major drawbacks in the system. As stated by the authors, the algorithm sometimes concludes closed eyes as open eyes owing to vertical rotation or lowering of the face. Secondly, the algorithm is highly dependent on illumination conditions. Finally, the authors have not considered the vibration of the car. Kumar *et al.* [33] proposed a heuristic approach for detection of eyes in close-up images. The algorithm is a three-stage process – thresholding in HSV color space, quantify spatially connected regions, followed by use of projection functions to localize the eye regions.

**Table 1-1: Summary of popular eye detection methods**

| Method | Author |
|---|---|
| Template matching | (Eriksson and Papanikolopoulos 1998) |
| Gabor wavelets | (Sirohey and Rosenfeld 2001) |
| Support Vector Machines | (Liu, Xu and Fujimura 2002) |
| Projection Functions | (Zhou and Geng 2003) |
| Principal Component Analysis | (Wang, et al. 2005) |
| Combined binary edge and intensity information | (Song, Chia and Liub 2006) |
| Pixel to edge information | (Asteriadis, et al. 2006) |
| Geometrical information | (Asteriadis, et al. 2006) |
| Neural Networks | (D'Orazio, et al. 2007) |
| Classifiers based on Haar-like features | (Hadid, et al. 2007) |
| Camshift prediction in YCbCr plane | (Hong and Qin 2007) |
| Skin color segmentation | (Liying and Haoxiang 2008) |

## 1.5 Eye Movements

The various types of eye movements in human beings. Of them, saccade, smooth pursuit and vergence have been widely studied in cognitive sciences. A saccade is a rapid and simultaneous movement of both the eyes in the same direction. It is the transition of the eye balls from one target fixation to another target fixation. In smooth pursuit, the eyes closely follow some moving object. Vergence is the movement of both eyes in opposite direction. The two most common types of vergence are convergence and divergence. Literature [2] suggests that peak saccadic velocity and saccadic duration bear a relationship with the alertness level.

## 1.6 Saccade and Alertness

Several approaches have been reported which establish the relationship of eye saccade with alertness level and to ascertain which saccadic parameter relate well with alertness. In [13], Inchingolo *et al.* studied the saccadic amplitude - peak velocity and saccadic amplitude - duration characteristics experimentally. The authors applied analog low-pass filtering to the recorded eye movement and then digitized it. They designed algorithms for computing the iris velocity, which are applicable for 200 Hz-sampled saccades. They validated their work using the Electro-Oculo-Gram (EOG) signal. Nevertheless, analysis was not carried out with respect to the level of alertness. In [34], the authors have investigated saccades along with blink and fixation pauses, and found out that blink frequency, saccadic velocity and frequency of fixations varies with change in alertness levels. They have employed EOG based methods to capture the eye movements. They conducted the experiment on 25 subjects. In [35], Deubel *et al.* investigated the relationship between visual attention and saccadic eye movements. The subjects were asked to distinguish between 'E' and '3' with surrounding distracter. They showed that the discrimination is best when the eye is directed towards the target. Nevertheless, estimation of saccadic parameters has not been discussed in the work. Ueno *et al.* [16] developed a system to assess human alertness and to alert the subject with acoustic stimulation on the basis of the dynamic characteristics of saccadic eye movement. The approach is model-based with a lot of approximations which do not emulate practical situations. The system has also not been tested in real-time.

## 1.7 Saccadic Parameter Estimation

According to literature [36], the popular methods for recording eye movements include the use of scleral search coils, EOG signals and video of eye image sequences. In the scleral coil based method, a magnetic coil is attached to the eye. A voltage signal which depends on the eye position will be generated in the coil and is captured. The major issue with scleral coil based technique is that attaching the coil on the sclera will cause high discomfort to the subject. In EOG based method, electrodes are carefully placed on the forehead as well as sides of the eye as shown in Fig. 1-1. It is possible to separately record horizontal and vertical movements using EOG. Nevertheless, the recordings are affected by metabolic changes in the eye [40]. It is reported in [37] of not being a reliable method for quantitative measurement, particularly of medium and large saccades. In ocular image based methods, active infra-red illumination is popularly used for locating the pupil center [37]. However, active infra-red illumination may cause harm to the subject on sustained exposure. The pupil detection methods working on passive illumination show limited accuracies in locating the pupil centre. Gupta *et al.* [3] have proposed a method for image-based saccadic parameter estimation in passive illumination using Form Factor based method.

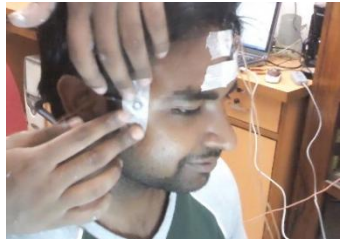

**Fig. 1-1 EOG electrode placements**

## 1.8 Pupil/Iris Centre Detection – A Review

It is apparent that for image based estimation of saccadic eye movements, fast and accurate detection of pupil/iris center is necessary. Pupil detection involves use of active illumination techniques to make the pupil region distinguishable from the iris. Several approaches have been reported for detecting and tracking the pupil using active illumination. In [38], Ebisawa proposed a new method of pupil detection for tracking the eye gaze. He used Bright Pupil - Dark Pupil (BPDP) method to detect the pupil. He discussed about removal of eye glass reflection light for subjects wearing spectacles. In [39], Morimoto developed a fast, robust and low cost pupil detection method using two Near Infra-Red (NIR) multiplexed light sources which are synchronized with camera acquisition speed. Once the pupil got located using the BPDP method, they tracked it using image difference having a threshold. The major limitation in BPDP method [38], [39] is that bright pupil method works well only in dark while dark pupil method works well only under bright light conditions. Moreover, sustained exposure to active illumination may be harmful for the eyes [40].

Considering the limitations of active illumination, we have considered the use of passive illumination for iris detection. Several approaches for iris detection under passive illumination has been reported in literature. Toennies *et al.* [41] discussed the feasibility of Circular Hough Transform based Iris localization for real-time applications. They obtained an iris radius of 50 pixels with a resolution of 100μm/pixel keeping the camera at a distance of 1m from the eye with a speed of 20.16 fps on an average on a 266 MHz Pentium II PC. The drawback is that performance the Hough Transform under varying illumination conditions was poor such as glints occurring on the image of eye surfaces. In [42], Kawaguchi *et al.* detected iris using combination of pixel intensity and edge information. They obtained a success rate of 95.3% on an average. However, they performed experiments on subjects devoid of glasses. Hansen *et al.* [31] proposed a log likelihood-ratio function of foreground and background models in a particle filter-based eye tracking framework. They fused key information from BPDP and their corresponding subtractive image into one single observation model. They obtained a detection rate of 91.17%. Valenti *et al.* [43] have performed eye centre location and tracking under passive illumination using isophotes of eye images. They have tested their algorithm robustness to changes in illumination and pose, using the BioID and the Yale Face B databases, respectively. The algorithm is reported to be robust against linear lighting changes, in-plane rotation and has low computational cost. Gupta *et al.* [3] proposed new features, called Form Factor to detect the iris.

## 1.9 Research Issues

From the above cited literature review, the issues pertaining to the accurate estimation of PERCLOS and saccadic parameters for real time applications are as follows:

- Improving the frame rate within real-time constraints
- Variation of illumination
- Face rotation
- Noise due to vibration
- Removal of glint from eye surface image for iris/pupil centre detection
- Lack of video databases of development and testing of algorithms

## 1.10 Contributions of the thesis

The major contributions in this thesis are as follows:

- Creation of five sets of video databases each pertaining to a particular application
- Development of a real-time framework for accurate estimation of PERCLOS suited for automotive driving environment
- Design of an embedded platform for monitoring the loss of attention in automotive drivers
- Development of an algorithm for estimating SR
- Establishment of Correlation between EOG and image based saccade measurement
- Development of an algorithm for detection of spectacles

## 1.11 Organization of the thesis

The thesis is organized as follows. Chapter 2 discusses about experiment design and database creation. Chapter 3 describes with the real time algorithm for estimating PERCLOS, with a case study for automotive driving. Chapter 4 presents a correlation study of EOG signals with image based estimation of eye saccadic parameters. Chapter 5 describes the algorithm for SR estimation. Chapter 6 reports an approach to detect the presence of spectacles in face images. Chapter 7 concludes the thesis with a future conception of a combined PERCLOS and saccade based system.

# Chapter 2. Design of Experiments

A major requirement for development and testing of algorithms to address the problem at hand is the availability of suitable databases customized for the particular application. In this work, five sets of video databases have been created, each corresponding to addressing a particular problem. The databases along with their purposes are listed below:

**Table 2-1: Database Descriptions**

| Database Number | Database Content | Purpose |
|---|---|---|
| Database I. | Human face | Effect of speed and accuracy of face detection with down-sampling |
| Database II. | Facial Images of drivers on board | Variation of illumination during day driving |
| Database III. | Near-infrared illumination | Facial feature detection during night driving |
| Database IV. | High speed eye movement videos | Estimating eye movement parameters |
| Database V. | Facial images containing as well as devoid of spectacles | Detection of spectacles |

For all the database creation, prior consent forms have been signed by the subjects. In this chapter, the experiment designs are explained in details.

## 2.1 Database I: Database of human faces for studying effect of down sampling

This database is a set of videos of face and non-face images for the studying the effect of down sampling on Haar-classifier based face detection algorithm with respect to speed and accuracy. Twenty subjects are selected in the age group of 20 to 35 years. Videos of facial and non-facial images are recorded in complex background under well-lit laboratory conditions at 30 fps at a resolution of $640 \times 480$ pixels using a USB camera and stored in AVI format (DivX codec). The subject was placed at a distance of 50 cm from the screen and the camera was mounted on top of the screen. Fig. 2-1 shows some sample images extracted from the videos. The subject was asked to move his head following an animated red circle projected on a screen. The positioning of the red circles is shown in Fig. 2-2. The circles followed a particular pattern which was repeated twice. In the first cycle, the subject was asked to follow the pattern with open eyes. In the next cycle, he/she was asked to repeat the same but with closed eyes. Next, the subject was asked to fake expressions such as smile, laugh, angry, fear, disgust etc. Finally, he/she was asked to move away from the camera to record some non-face images.

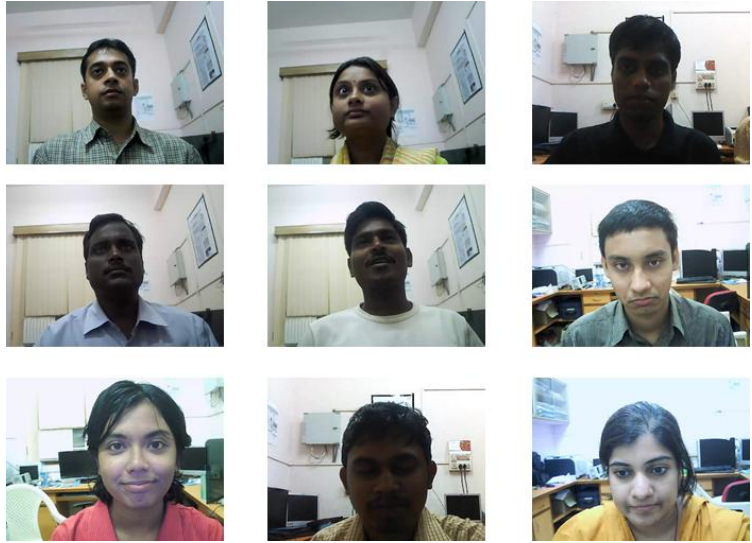

**Fig. 2-1 Sample images from Database - I**

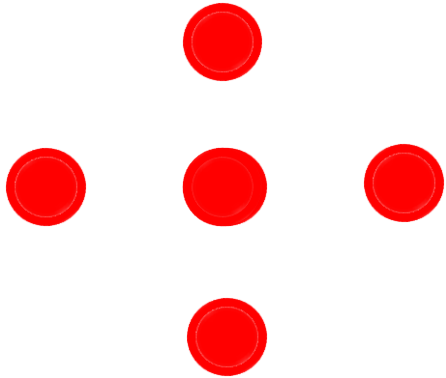

**Fig. 2-2 Positioning of the circles**

## 2.2 Database II: Database of human faces captured on-board during day driving

This database has been created to evaluate the performance of the face and eye detection algorithms under different on-board conditions of the day against variation in illumination and extreme lighting conditions. The database consists of videos of six drivers. The drivers were all male in the age group of 25 to 35 years age. Some images taken from the database are shown in Fig. 2-3. A USB camera was placed on the dashboard and was connected to a laptop for recording the videos. The driver was asked to drive on rough roads as well as on highways. The recording was carried out in the midday during the period of 11:00 am to 1:00 pm. The recordings were performed at 640 × 480 pixels resolution at 30 frames per second (fps) using DivX codec in AVI format.

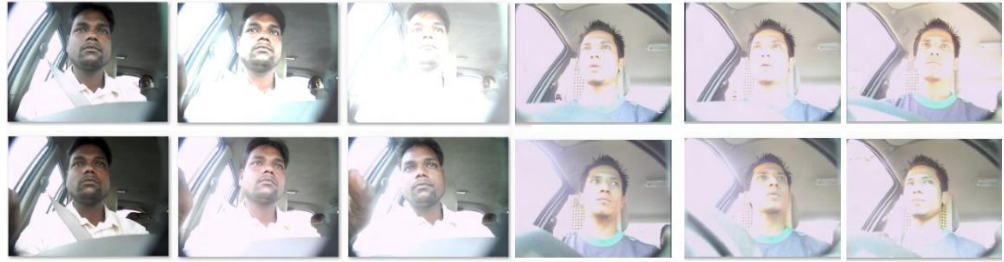

**Fig. 2-3 Sample images taken from Database – II**

## 2.3 Database III: Face Database under Near Infra-Red (NIR) illumination

The objective of this database was to evaluate the performance of the face and eye detection algorithms for night driving, in which the images were captured under NIR illumination. The experiments were conducted in a dark laboratory with all sources of visible lights closed during the recording. The subjects were given video as well as verbal instructions regarding the tasks they had to perform. Sixty subjects have been selected of which 40 are male and 20 are female. The subjects are in the age-group of 20 to 40 years. The recording time for each subject was eight minutes on an average. The recordings were performed at 640 × 480 pixels resolution at 30 fps using DivX codec in AVI format. The videos were captured using a USB camera with IR filter removed. An NIR lighting system made up of Gallium Arsenide LEDs is prepared taking into consideration the safety illumination limits. Audio instructions which instructed the subjects regarding the movement of the face, were provided during the video recording. An arrangement was made to prevent the illumination from the laptop screen. The lighting system consisted of three arrays of Ga-As NIR LEDs. The LEDs were mounted on a movable stand which provided both horizontal as well as vertical movements. The lighting intensity was varied by changing the relative position of the LEDs as well as changing the power supply. Fig. 2-4 shows some sample images taken from this database. The details of the complete database creation has been published in [44].

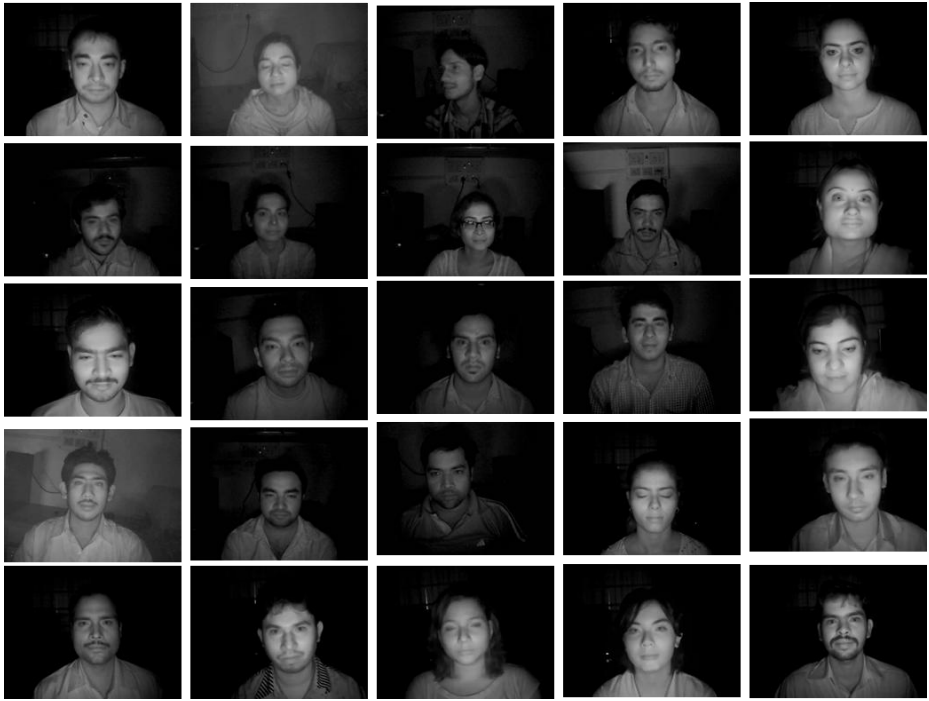

**Fig. 2-4 Sample images from Database III**

## 2.4 Database IV: Database of eye saccades

The purpose of this database is to evaluate the performance of algorithms for saccadic parameter estimation. EOG data were also collected along with the video data, for correlation studies. This database consists of eye saccade videos captured at 420 fps at $224 \times 168$ resolution using a high speed camera. The experiment was conducted under laboratory conditions with special illumination to account for the low exposure time of the high speed imaging. Twenty subjects in the age group of 20 to 35 years participated voluntarily for the experiment. The camera was focused at the eye and the recordings were captured synchronously with that of EOG data. The camera was placed 20 cm from the subject's eyes. A tungsten filament bulb was used for illumination. The reason for external illumination is that, because of high frame rate acquisition, the exposure time becomes low, which results in poor illumination in the images thereby causing loss of information. The subject was asked to follow a red target on a screen which flashed in such a manner so as to make the subject execute full horizontal visually triggered saccades. Fig. 2-5 shows some images of dataset. A total of 24 full horizontal saccades were recorded for each subjects.

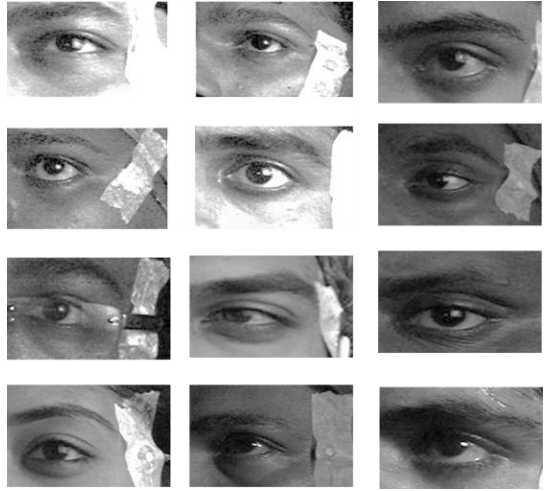

**Fig. 2-5 Sample images taken from Database IV**

## 2.5 Database V: Database of faces containing as well as devoid of spectacles

This database has been created for the identification of the presence of spectacles in a given face image. The database consists of human faces consisting of as well as devoid of spectacles. Twenty subjects in the age group of 20-35 years participated in the experiment. Both male and female subjects were selected and there were asked to wear different kinds of spectacles during the experiment. The experiment was conducted under laboratory conditions with a variation in illumination and face pose. The subjects were asked to rotate their faces in a predefined systematic fashion. The same animation which was used for the creation of Database I has been used here. The recordings were performed at $640 \times 480$ pixels resolution at 30 fps. Some sample images of different subjects wearing spectacles are shown in Fig. 2-6. In Fig. 2-7, some sample images of different subjects are shown with faces devoid of spectacles.

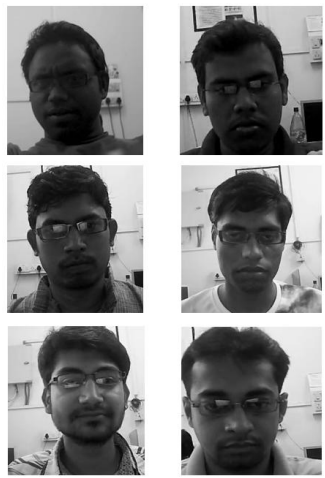

**Fig. 2-6 Sample images from the Database V (wearing spectacles)**

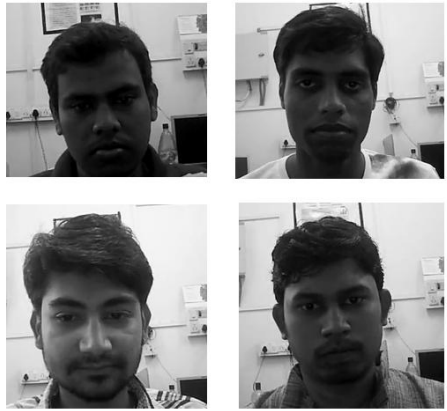

**Fig. 2-7 Sample images from the Database V (Devoid of spectacles)**

## 2.6 Conclusion

This chapter discusses about the creation of video databases. Five sets of databases have been created which has been used later in this work for development as well as performance evaluation of the algorithms. The databases will be helpful for future research as well. The first database is that of human faces for studying the effects of downscaling the frames with respect to speed and accuracy of Haar classifier based face detection algorithm. The second database is that of faces of drivers taken on-board during actual driving condition under different illumination levels during day driving. The third video database is that of faces taken under NIR illumination. The purpose of this database is to test the algorithm for night driving conditions where the face of the driver will be illuminated with NIR LEDs as discussed in the next chapter. The fourth database is that of high speed eye saccade videos for saccadic parameter estimation. The database has also been used for correlation of image based estimation with that of EOG signal based method. The fifth database is that of videos of facial images, consisting of as well as devoid of spectacles. All the databases will be available upon email at anirban1828@gmail.com.

# Chapter 3. Real-time Computation of PERCLOS

This chapter describes about the real-time algorithm for the image based estimation of PERCLOS. PERCLOS may be defined as the percentage of time where the eyelid remains at least 80% closed over the pupil. In the present framework for PERCLOS estimation, face is first detected in real-time, followed by selection of a Region Of Interest (ROI) in which the eye is localized and then classified as open or closed. Here closed eye denotes at least 80% spatially closed over the pupil. Finally, the PERCLOS value is computed as defined in 1-1. A 66.67% (two-third) overlapping time window of three minutes duration is used to compute the PERCLOS value for every minute. A case study for estimating PERCLOS for a human driver in automotive environment is presented.

## 3.1 Real-time Face Detection

Several algorithms for face detection have been reported in literature which are based on neural networks [45], Support Vector Machines (SVM) [46], Example-based learning [47], Eigen faces [48], Fischer faces [49], Template Matching [50], Haar-like features [51]. Among these methods, the use of Haar-like features have been reported to be the most robust real-time method for face detection [51]. In this thesis, the use of Template Matching, PCA (based on Eigen faces) and Haar-like features have been tested for face detection and Haar-like features have been finally selected for face detection owing to its better performance in the particular context.

### 3.1.1. Face Detection using Template Matching

This section describes the detection of face using Template Matching [54]. The first step in this algorithm is to create the face templates for training. One hundred grayscale face templates of size 200×140 pixels is selected randomly from Database I ensuring variation in the training set. Now, the templates are matched with an input frame in real-time successively. A threshold of 0.8 per unit is set which is obtained empirically. A gray template image, $w(x, y)$ of size $J \times K$ is examined with a test image $f(x, y)$ of size $M \times N$ with $J \leq M$ and $K \leq N$, where $x = 0,1, \ldots M - 1; y = 0,1, \ldots N - 1$, $\bar{w}$ is the average value of $w$ and $\bar{f}$ is the average value of that window where $f$ is covered by $w$. The normalized correlation coefficient $\gamma$ is used to match the template in each search window. The value of $\gamma$ is computed using:

$$\gamma(x,y) = \frac{\sum_s \sum_t [f(s,t) - \bar{f}(s,t)] \, [w(x+s,y+t) - \bar{w}]}{\{\sum_s \sum_t [f(s,t) - \bar{f}(s,t)]^2 \sum_s \sum_t [w(x+s,y+t) - \bar{w}]^2\}^{\frac{1}{2}}} \qquad \text{3-1}$$

$\gamma$ indicates the similarity between the template and the test image. A spatial window having 25% overlap in both horizontal and vertical directions has been used. If the $\gamma$ value is at least 80% similar to any of the templates, then the test window is considered as a face in the algorithm.

The above algorithm has been implemented on two embedded devices as stated under –

- a Smart Camera having 633 MHz clock speed, 128 MB RAM
- a Single Board Computer (SBC) having 1.6 GHz clock speed, 128 MB RAM with a USB camera

The method was found to have a hit rate of approximately 85.5% for face detection but a high false alarm rate of about 30.2% on Database I. A speed of 2.5 fps was achieved on the SBC and 1.5 fps on the Smart Camera using this method. Hence, this method was rejected owing to high false alarm rate and poor runtime performance.

### 3.1.2. Face Detection using PCA

This section describes real-time PCA [52] based face detection. The same face templates used for Template Matching were used in this method. The training algorithm is stated below:

**Table 3-1: Training Algorithm**

| |
|---|
| 1. Face images $I_1$, $I_2$, $I_3$......$I_{100}$, each of dimension 200×140 pixels are collected |
| 2. Every image $I_i$ forms a vector $\Gamma_i$ (of dimension 28000× 1) |
| 3. The average face vector is computed using |
| $$\psi = \frac{1}{P} \sum_{i=1}^{P} \Gamma_i \, , where \; P = 100$$ |
| 4. The mean face is subtracted from each image vector $\Gamma_i$ |
| $$\phi_i = \Gamma_i - \psi$$ |
| 5. The estimated covariance matrix, $C$ is given by: |
| $$C = \frac{1}{P} \sum \phi_n \phi_n^T = AA^T$$ |
| Where $A = [\phi_1, \, \phi_2... \, \phi_P]$ (28000×$P$ matrix) |
| As $C$ is very large ($P \ll 28000$), $A^T A$ ($P \times P$) is computed |
| 6. The eigenvectors $v_i$ of $A^T A$ are computed using |
| $$\sigma_i u_i = A v_i$$ |

Using the equation above Eigenvectors $u_i$ of $AA^T$ are obtained

7. Only $K$ eigenvectors are retained corresponding to the $K$ largest eigen values. These $K$ eigenvectors are the Eigen faces corresponding to the set of the face images

8. These $K$ eigenvectors are normalized

The value of K was selected as 15 empirically. The face detection is carried out by the method as given below:

**Table 3-2: PCA based Face Detection Algorithm**

1. 25% overlapping windows, $\Gamma$ of size 200×140 are used for testing. Then, $\phi$ is obtained from the mean image $\psi$ as
$$\phi = \Gamma - \psi$$

2. Compute
$$\widehat{\phi} = \sum_1^k w_i \, u_i$$

3. The error, $e$ is computed as
$$e = \left\| \phi - \widehat{\phi_j} \right\|$$

4. The window corresponding to the minimum error, $e$ is selected as face

The above algorithm has been implemented on the same Smart Camera and SBC. The method was found to have a true positive rate of about 74.5% and a false alarm rate of about 26.2% on Database I. A speed of 3 fps was achieved on the SBC and 1 fps on the Smart Camera using this method. The major issue found with this method was that the performance of the algorithm was prone to variation with distance from the camera.

### 3.1.3. Face detection using Haar-like features

Haar-like features [51] are rectangular features in an image used in object detection. Viola *et al.* [20] devised an algorithm to rapidly detect any object, using AdaBoost classifier cascades that are based on Haar-like features. Fig. 3-1 shows two-rectangle and three-rectangle Haar-like features. The sum of the pixels which lie within the white rectangles is subtracted from the sum of pixels in the black rectangles.

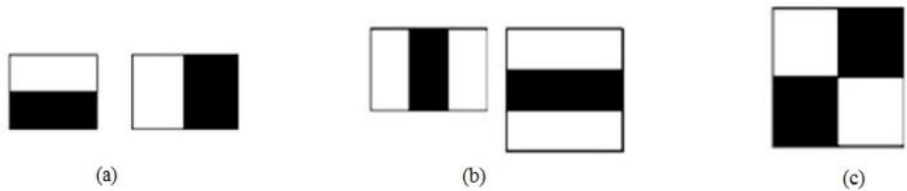

(a)        (b)        (c)

**Fig. 3-1 (a) two rectangle features (b) three rectangle features (c) four rectangle features**

The value of a two-rectangle feature is the difference between the sum of the pixels within two rectangular regions of same size and shape which are horizontally or vertically adjacent. A three-rectangle feature computes the sum within two outside rectangles subtracted from the sum in a center rectangle. A four-rectangle feature computes the difference between diagonal pairs of rectangles.

For fast computation of rectangular Haar-like features, integral images are computed. Each element of the integral image contains the sum of all pixels located in the up-left region of the original image. Using this, the sum of rectangular areas in the image is computed at any position or scale. The integral image, $I(x, y)$ is computed efficiently in a single pass over the image using:

$$I(x, y) = i(x, y) + I(x - 1, y) + I(x, y - 1) - I(x - 1, y - 1) \qquad 3\text{-}2$$

where, $I(x, y) = \sum_{\substack{x' \le x \\ y' \le y}} i(x', y')$ and $i(x, y)$ is the intensity at point $(x, y)$.

The face detector scans the input image at 11 scales, starting at the base scale of 24×24 pixels and with each scale 1.25 times of the previous. There are 45,396 rectangle features [52] associated with each image sub-window, which is far larger than the number of pixels. Even though each feature can be computed very efficiently, computing the complete set is cumbersome. Viola *et al.* [52] stated that a very small number of these features can be combined to form an effective classifier. The main challenge is to find these selected features. Voila *et al.* [52] have used Gentle Adaptive Boosting (AdaBoost) for the purpose. To boost the weak learner, it is called upon to solve a sequence of learning problems. For each feature, the weak learner determines the optimal threshold classification function, such that the minimum number of examples is misclassified. A weak classifier $h_j(x)$ consists of a feature $f_j$, a threshold $\theta_j$ and a parity $p_j$ indicating the direction of the inequality sign as given in 3-3.

$$h_j(x) = \begin{cases} 1 \ if \ p_j f_j(x) < p_j \theta_j \\ 0 \qquad otherwise \end{cases} \qquad 3\text{-}3$$

Here $x$ is a 24×24 pixel sub-window of an image. The learning algorithm is shown in Table 3-3. $T$ numbers of hypotheses are constructed, each using a single feature. The final hypothesis is a weighted linear combination of these $T$ hypotheses where the weights are inversely proportional to the training errors.

The cascading of the weak classifiers to obtain a strong classifier is depicted in Fig. 3-2. The structure of the cascade reveals that within any single image, a majority of sub-windows are negative. So the cascade attempts to reject as many negatives as possible at the earliest possible stage.

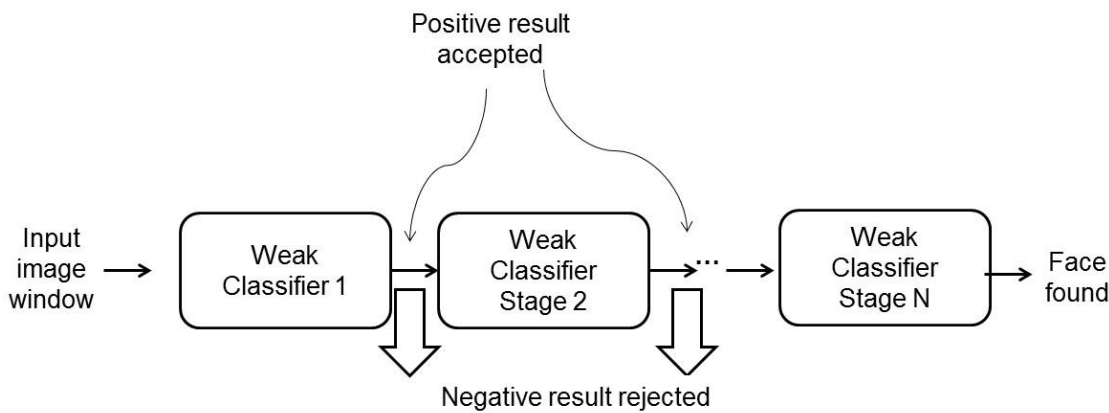

**Fig. 3-2 A series of classifiers is applied to every sub window**

Given a trained cascade of classifiers, the false positive rate,$F$ on the training set of images, is

$$F = \prod_{i=1}^{K} f_i \qquad 3\text{-}4$$

where $K$ is the number of classifiers, and $f_i$ is the false positive rate of the $i^{th}$ classifier on the examples that get through it. The detection rate $D$ with $d_i$ as the detection rate of the $i^{th}$ classifier, on the training set, is

$$D = \prod_{i=1}^{K} d_i \qquad 3\text{-}5$$

The expected no. of features which are evaluated, for $n_i$ and $p_i$ being the no. of features and positive rate for $i^{th}$ classifier respectively, is

$$N = n_o + \sum_{i=1}^{K} (n_i \prod_{j<i} p_j) \qquad 3\text{-}6$$

**Table 3-3: Algorithm for Gentle AdaBoost**

| Algorithm: Gentle Adaptive Boosting |
|---|
| 1. Training images $(x_1, y_1), \ldots, (x_n, y_n)$ are provided with $y_i = 0$ for negative image and 1 for positive image. |
| 2. Weights $w_{1,i} = \frac{1}{2m}, \frac{1}{2l}$ for $y_i = 0,1$ respectively are initialized, where $m$ = no. of negatives and $l$ = no. of positives. |
| 3. For $t = 1, \ldots, T$ |
|      i. The weights, $\frac{w_{t,i}}{\sum_{j=1}^{n} w_{t,j}} = W_{t,j}$ are normalized so that $w_t$ is a probability distribution |
|      ii. For each feature $j$, a classifier $h_j$ is trained which is restricted to using a single feature. The error is evaluated with respect to $w_t$, $\epsilon_t = \sum_i w_i \lvert h_j(x_i) - y_i \rvert$ |
|      iii. The classifier,$h_t$ is selected with the lowest error $\epsilon_t$ |
|      iv. The weights $w_{t+1,i} = w_{t,i}\beta_t^{1-e_i}$ are updated where |

$$e_i = \begin{cases} 0 \ if \ x_i \ is \ classified \ correctly \\ 1 \qquad\qquad otherwise \end{cases}$$

$$\beta_t = \frac{\epsilon_t}{1 - \epsilon_t}$$

4. The final strong classifier is $h(x) = \begin{cases} 1 \ if \ \sum_{t=1}^{T} \alpha_t h_t(x) \geq \frac{1}{2}\sum_{t=1}^{T} \alpha_t \\ 0 \qquad\qquad\qquad otherwise \end{cases}$

Where $\alpha_t = log \frac{1}{\beta_t}$

In the next section, it is shown how the training parameters affect the performance of the classifier.

### 3.1.4. Training of Classifiers based on Haar-like Features

The training of the Haar based classifier for face detection requires proper selection of training parameters for optimum results. The training includes several steps as described in [53]. It is required to collect positive images that contain only objects of interest (faces in the present context) as well as the negative images which are devoid of the object of interest to build the classifier. The following steps are taken to obtain the final classifier.

**Table 3-4: Steps for Haar Training**

Algorithm: Haar training

1. Collection of training images containing faces in different orientation under varying illumination.

2. Positive and negative images were generated along with corresponding information texts files like 'positive.txt' and 'negative.txt'. The text files contain the coordinates of the positive and negative training samples along with their names in a specified order. The name of the image file contains the following format.

   $< File \ Name >< x >< y >< width >< height > .jpg$

   where *x, y, width*, and *height* define the object bounding rectangle.

3. A vector (.vec) file is obtained from the text files generated in the previous step. The vector file contains compact information of positive instances of objects and the negative image such as background.

4. The classifiers are trained using the vector file.

5. Once the Haar training is complete, the output is stored in the specified directory in the form of text files. The final step of the process is to convert these files into a single xml file. This xml file was the final classifier.

### 3.1.1. Effect of training parameters on face detection

Before using the classifier, it is necessary to choose an optimum one based on its performance. Hence, it is essential to examine the detection performance of the algorithm considering a few Haar-like features and including varieties of human faces with different orientations. Towards this goal, the issues related to the selection of design parameter values to improve the detection rate are critically reviewed and tested. Experiment is carried out to record faces with different orientations under varying illumination conditions in laboratory. Five classifier sets are developed by varying two design parameters (a) maximum angles and (b) maximum deviation during the training process. These are tested on eight test image sets generated from the recorded images which were different from the training set. Receiver Operating Characteristics (ROC) curves are generated for each of the classifiers and Area Under the Curve (AUC) of these ROC's are computed to compare their performances. Two datasets of images are prepared for the training. One set contains the faces (positive images) or the other set contains non-faces (negative images). The set of positive images are selected from the dataset created as described in the thesis of [54]. Information files are created to contain the information about the coordinates of the objects in the original dataset. The original positive dataset is made of images taking into consideration the variance between different people, including race, gender, and age to produce a very robust detector.

*1) Training Data Set:* The training database is formed from these extracted frames selected from the video database created in the thesis of [54]. A few examples of positive training images are shown in Fig. 3-3.

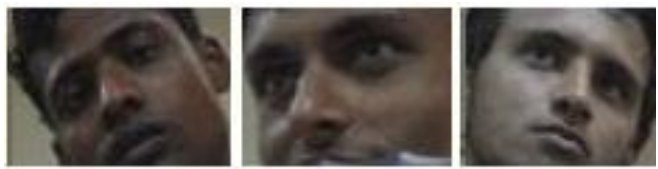

**Fig. 3-3 Sample Positive Images (Faces of human subjects)**

*2) Testing Data Set:* The test data set was formed from the earlier extracted frames excluding the images contained in the training database. Some non-face images are also added with this test data set to account for false positives and false negatives. All the images in this database are kept at same resolutions. Subsequently, this data set is divided into eight groups corresponding to eight subjects. Each of these groups contains 115 images of both faces and non-faces. Each of these images within a group is manually identified and noted for computation of the ROC curve for each classifier.

*3) Training Parameters:* In this work, we have varied the 'Maximum Angle' (MA) and 'Maximum Deviation' (MD), to study their effect on classification performance. MA is the maximum rotation angles of the features from the original training images along three different axes. MD is the maximal intensity deviation of pixels in foreground samples.

### 3.1.2. Classifier Design and Generation

The classifier sets are formed using the conventional Haar training process as given in Table 3-4. Five sets of classifiers are developed by varying the MA's and MD's. The classifier sets are defined in Table 3-5.

**Table 3-5: Definition of Classifier Sets**

| Classifier | M.A. | M.D. |
|------------|------|------|
| Set 1 | 30 | 100 |
| Set 2 | 45 | 80 |
| Set 3 | 45 | 100 |
| Set 4 | 45 | 120 |
| Set 5 | 60 | 100 |

### 3.1.3. ROC Curve and AUC Calculation

An ROC curve [55] is used to visualize the performance of a machine learning algorithm like the classifiers considered in this work. This is a mapping between true positive rate (tpr) and false positive rate (fpr). The points on the ROC curves computed from the test database are interpolated to construct a smooth average version of it for each of the classifier. The tpr and fpr are defined in 3-7 and 3-8 respectively in terms of true positives (tp), false positives (fp), false negatives (fn) and true negatives (tn).

$$\mathbf{tpr} = \frac{\mathbf{tp}}{\mathbf{tp+fn}} \qquad \qquad 3\text{-}7$$

$$\mathbf{fpr} = \frac{\mathbf{fp}}{\mathbf{fp+tn}} \qquad \qquad 3\text{-}8$$

The performance of each classifier is evaluated using the ROC curves. The Area Under the Curve (AUC) is used to indicate the accuracy of the classifier.

### 3.1.4. Results and Discussions

The number of tn, tp, fp and fn are computed. Fig. 3-4 shows some detection results using Classifier 1. Fig. 3-5 gives the ROC comparisons for different classifier sets.

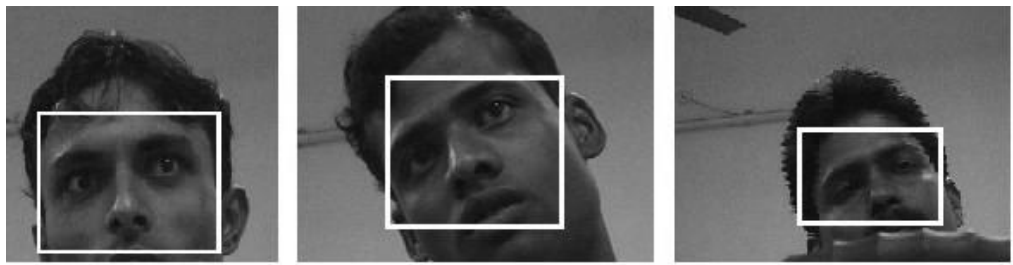

**Fig. 3-4 Examples of True Positives using Classifier Set 1**

From Table 3-6, it is found that the AUC of each classifier set is above 80% which indicates that the classifier sets have a high hit rate with good accuracy and low false alarming. Set 3 gives the highest AUC which is an indication that it is the best among these five classifier sets. The mean of the AUC is found to be 0.8313 with a standard deviation of ±0.0108. The mean value shows that all the classifier sets perform nicely on an average basis. It can be observed from Table 3-6 that in Sets 1, 3, and 5, that the performance of the sets reaches maximum with a moderate value of either MA or MD. This reveals that the MA and MD may not be selected arbitrarily high or low. Even though, a moderate value of them may produce classifier with reasonable performance.

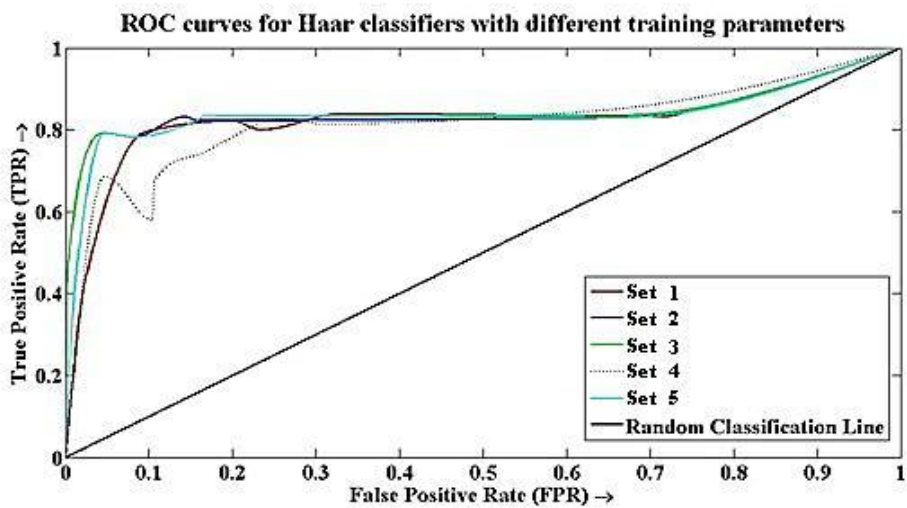

**Fig. 3-5 ROC comparison for different classifier sets**

**Table 3-6 AUC values for each classifier set**

| Classifier Sets | AUC |
|---|---|
| Set 1 | 0.8362 |
| Set 2 | 0.8167 |
| Set 3 | 0.8447 |
| Set 4 | 0.834 |
| Set 5 | 0.8249 |

### 3.1.5. Real-time Framework for Face and Eye Detection

Once the classifier is tested offline with proper tuning of parameters, it is deployed to detect the face from images from a USB camera in real time. The detection rate was found to be 3 fps, which is low for accurately estimating the state of the eye. The subsequent sections discuss about the real-time framework designed to improve the runtime performance of the algorithm without compromising on accuracy.

### 3.1.6. Improving the real-time performance

The integral image computation time is variable and depends on the number of image pixels. The incoming frame of size $N \times M$ is down sampled to perform the detection at higher frames rates to improve the real- time performance. Bicubic interpolation, being more accurate that Bilinear and nearest neighbor interpolations is used for down sampling [56]. The number of pixels in the new image is now reduced. We define the down sampling Scale Factor (SF) as:

$$SF = \frac{no. of\ vertical\ pixels\ in\ original\ frame}{no. of\ vertical\ pixels\ in\ down\ sampled\ frame} = \frac{no. of\ horizontal\ pixels\ in\ original\ frame}{no. of\ horizontal\ pixels\ in\ down\ sampled\ frame} \quad 3\text{-}9$$

For an SF of $k$, the number of pixels is reduced to $\frac{N}{k} \times \frac{M}{k}$. The time taken to compute the integral image is reduced as the number of search windows is reduced.

### 3.1.7. Extraction and Remapping of ROI co-ordinates

The following table describes the framework of face and eye detection for estimating PERCLOS.

**Table 3-7: Scheme of face and eye detection**

| Algorithm: Framework of face and eye detection |
|---|
| 1) The image coming from the camera is scaled down by a factor of $k$. The original image is stored before it is scaled down. |
| 2) Face is detected in the down sampled image using the Haar classifier for face. |
| 3) The co-ordinates of the face are obtained from the Haar classifier in the down sampled frame. |
| 4) An ROI is selected from the detected facial image, based on the morphology of the face, as it is evident that eye occupy the upper half of the face |
| 5) ROI co-ordinates are extracted from the detected frame and remapped on to the original frame as shown in Fig. 3-6. |

> 6) Eyes are then detected in the selected ROI, using algorithms explained later in the thesis.

This method achieves higher detection rate primarily because of the low resolution face detection. The speed of the algorithm is dependent on the down sampling SF.

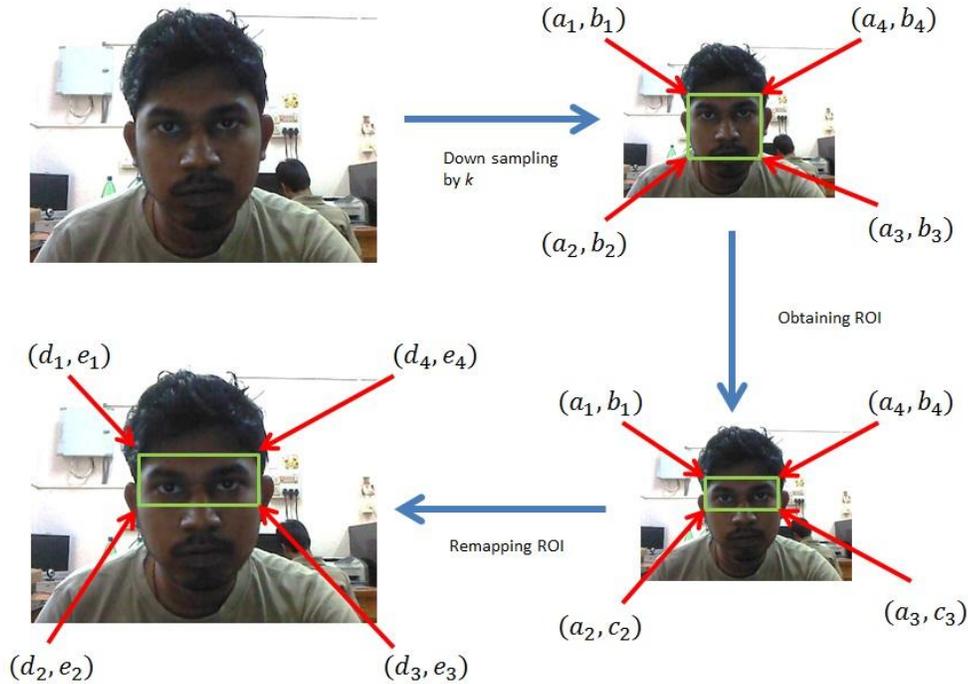

**Fig. 3-6 Scheme of eye detection**

In Fig. 3-6, $a_i's$ & $b_i's$ for $i = 1,2,3,4$ are obtained from the face detector based on Haar Classifier. ROI co-ordinates $c_2$ & $c_3$ are obtained as $c_2 = \frac{b_1 + b_2}{2}$ and $c_3 = \frac{b_3 + b_4}{2}$ respectively. The ROI co-ordinates $d_i's$ & $e_i's$ are obtained using relation in 3-10 and 3-11.

If $D = \begin{bmatrix} d_1 \\ d_2 \\ d_3 \\ d_4 \end{bmatrix}, E = \begin{bmatrix} e_1 \\ e_2 \\ e_3 \\ e_4 \end{bmatrix}, A = \begin{bmatrix} a_1 \\ a_2 \\ a_3 \\ a_4 \end{bmatrix}$ & $B = \begin{bmatrix} b_1 \\ c_2 \\ c_3 \\ b_4 \end{bmatrix}$, then

$$\boldsymbol{D = k \times A}$$ 3-10
$$\boldsymbol{E = k \times B}$$ 3-11

### 3.1.8. Choice of Optimum SF with respect to runtime and accuracy

As the SF increases, the speed is found to increase. A reduction in accuracy is observed, because of reduction in information in the downscaled frame. Hence an optimal SF is needed which gives a trade-off between speed and accuracy.

Database I has been used to observe the change in processing speed and accuracy with down sampling. The processing is carried out in a computer having Intel dual core processor 2 GHz clock, 2 GB of DDR2 RAM. The incoming frame is down sampled by SF's of 2, 4, 6, 8 and 10 successively. The processing speed for each video is noted down and average speed of all the subjects is plotted as shown in Fig. 3-7. The run-time versus SF plot reveals that the speed of operation increases almost linearly with SF.

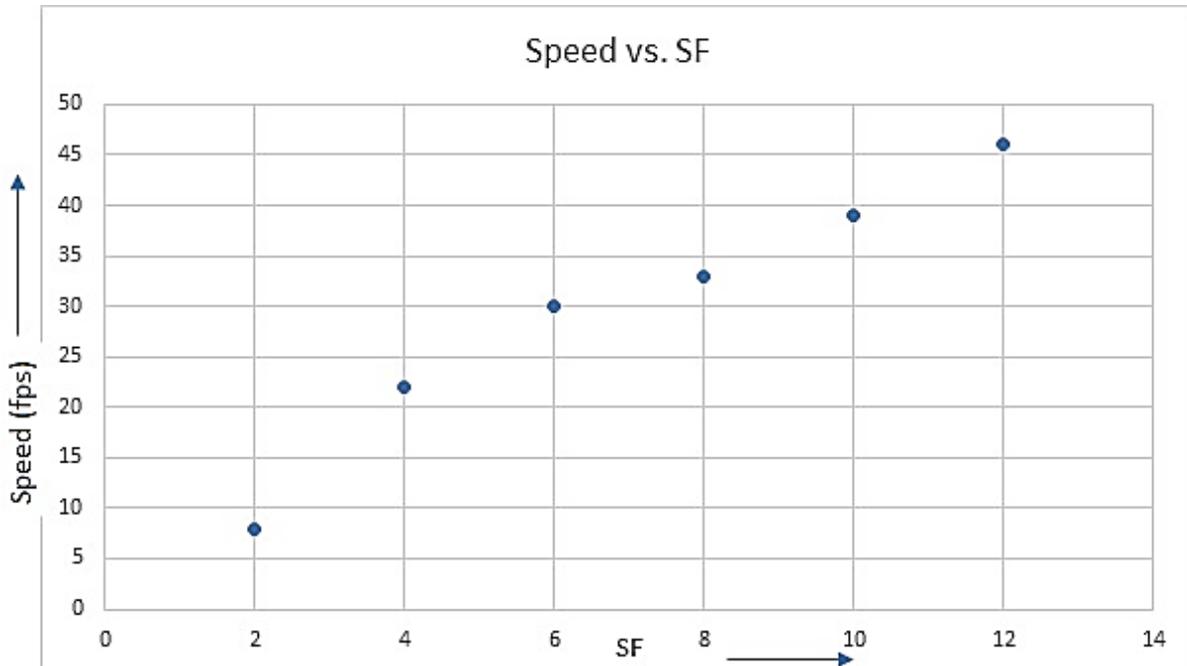

**Fig. 3-7 Speed vs. SF**

The accuracy of detection is obtained by plotting ROC curves and then calculating the AUC. The images are manually marked to take into account the presence of face to store the ground truth. Subsequently the algorithm for face and eye detection is executed on the same dataset. The detection results are then compared with the ground truth to obtain the number of tp, fp, tn and fn. The tpr and fpr are calculated using 3-7 and 3-8 respectively and ROC curve for each SF is obtained as shown in Fig. 3-8.

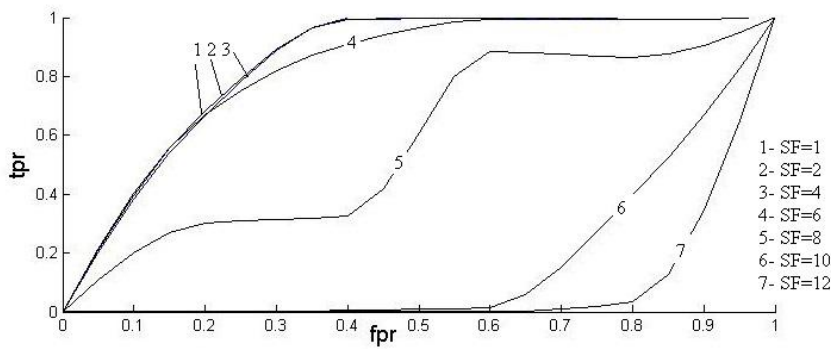

**Fig. 3-8 ROC comparisons for different SF's**

Table 3-8 shows the AUC values for each SF. It is observed that the AUC remains almost constant up to an SF of 4, and then decreases sharply after an SF of 6. From the observation, it is apparent that an SF somewhere in between 4 to 6 is expected to give optimum result. Hence, an SF of 5 is taken as the optimal one as a trade-off between run-time and accuracy.

**Table 3-8: AUC vs. SF**

| SF | Area Under the Curve (AUC) |
|----|----------------------------|
| 1  | 0.8460 |
| 2  | 0.8461 |
| 4  | 0.8493 |
| 6  | 0.8274 |
| 8  | 0.5790 |
| 10 | 0.1723 |
| 12 | 0.0833 |

### 3.2 In-plane Rotated Face detection

Detection accuracy is found to be low for a moderate amount of in-plane rotation of the face of the subject. In [57], an affine transformation based method has been used for the detection of in-plane rotated faces. The translation of the face is taken care of by the Haar classifier, as it performs a sliding window based search. The origin of the system is the first top-left pixel with co-ordinates (0,0). The rotation matrix $R$ is found from the angle of rotation $\theta$ of the image as shown in 3-12. If a point on the input image is $A = \begin{bmatrix} x \\ y \end{bmatrix}$ and the corresponding point on the affine transformed image is $B = \begin{bmatrix} x' \\ y' \end{bmatrix}$, then $B$ can be related to $A$ using 3-13.

$$R = \begin{bmatrix} cos\,\theta & -sin\,\theta \\ sin\,\theta & cos\,\theta \end{bmatrix} \qquad 3\text{-}12$$

$$A = RB \qquad 3\text{-}13$$

==The Haar classifier used for face detection was trained with inherent tilted Lienhart features for detection up to ± 15 degree. With affine transform, the additional in-plane rotation was taken care off.== The angles of rotation, $\theta$ are given values of $\pm 45^o$ and $\pm 30^o$ successively. After the face detection, the algorithm returns the ROI for eye detection in a de-rotated form. Fig. 3-9 shows tilted face detection using an affine transformation of the input image. The algorithm for detection of in-plane rotated faces is provided in Table 3-9.

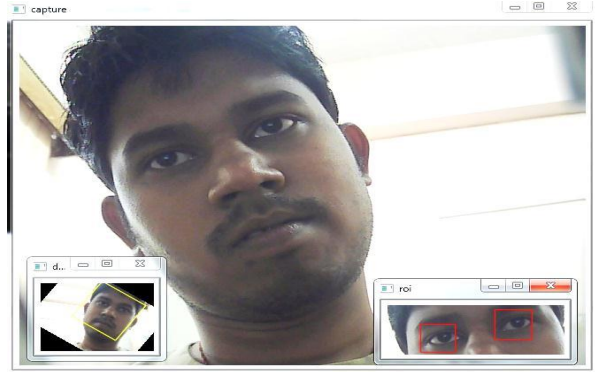

**Fig. 3-9 In-plane rotated face detection**
**Table 3-9: Algorithm for detection of in-plane rotated faces**

> 1) The Haar classifier checks for the presence of frontal faces in the input image.
>
> 2) If frontal face is found it follows the scheme as given in Table 3-9.
>
> 3) If frontal face is not found, we form a rotational matrix $R$ with $\theta = 30^o$, to transform the image using 3-13.
>
> 4) The Haar classifier searches for the face in the transformed image. If frontal face is found, it follows the scheme as given in Table 3-7.
>
> 5) If the frontal face is still not found, we form a rotational matrix $R$ with $\theta = -30^o, \& \pm 45^o$ successively to search for faces in the respective transformed frames.
>
> 6) If for none of the set values of $\theta$, the Haar classifier detects the face, the algorithm concludes that no face is found.

### 3.3 Compensation of effect of lighting conditions

The face detection accuracy was found to vary with illumination levels. However, for accurate estimation of PERCLOS for real scenarios such as on-board driving, the algorithm has to be illumination invariant. Hence compensation of the illumination level was necessary to achieve the goal. The compensation algorithm has to be used as a pre-processing step and hence should not be computationally cumbersome to affect the real-time performance. Several algorithms [58] [59] [60] [61] [62] [63] have been reported in literature which are almost illumination invariant. The works in [60] [61] [62] [63] are computationally intensive and hence is not considered as a pre-processing step in the present case. The work in [58] uses Bi-Histogram Equalization (BHE) of the pixel intensities to compensate for the varying illumination levels. The algorithm is accurate as well as computationally efficient and hence is well suited for the present application. In case the face is not detected even after the affine transformation, the face detection framework considers the problem as an illumination variation issue and performs BHE as a preprocessing step. In this method, the input image is decomposed into two sub-images. One of the sub-images is the set of samples less than or equal to the mean image intensity whereas the other one is the set of samples greater than the mean image. Histogram Equalization is used to equalize the sub-images independently based on their respective histograms under the constraint that the pixel intensities in the first subset are mapped onto the range from the minimum gray level to the input mean intensity while the pixel intensities in the second subset are mapped onto the range from the mean intensity to the maximum gray level. Hence, the resulting equalized sub-images are bounded by each other around the input mean. The mean intensity of the image is preserved which makes the method illumination invariant.

Mathematically representing the algorithm, let the input image $X$ be subdivided into two sub-images $X_L$ and $X_U$ based on the mean intensity image $X_m$ such that $X = X_L \cup X_U$, where

$$X_L = \{X(i,j)|X(i,j) \le X_m, \forall X(i,j) \in X\} \qquad \text{3-14}$$
$$X_U = \{X(i,j)|X(i,j) > X_m, \forall X(i,j) \in X\} \qquad \text{3-15}$$

Now, histogram equalization is carried out on images $X_L$ and $X_U$ independently. Fig. 3-10 shows the comparison of normal and corrected images of a driver under on-board conditions. The algorithm is evaluated against 200 images randomly selected from Database II. Using BHE, the face detection accuracy was found to improve from 92% to 94% approximately, without appreciable reduction in speed. There is scope of further improvement to compensate extreme illumination levels.

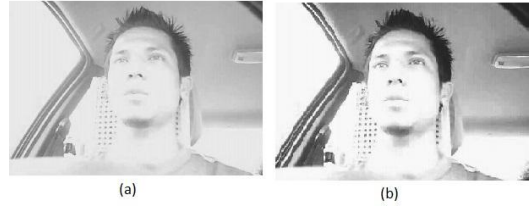

(a)          (b)

**Fig. 3-10 (a) Original and (b) Bi-Histogram Equalized images**

## 3.4  Real-time eye detection

Once the face is localized and ROI has been remapped, the next step is to detect the eyes in the ROI. Subsequently the detected eye is classified as open or closed to obtain the PERCLOS value. In this work the eye detection has been carried out by a PCA based algorithm during day whereas Block LBP features are used for eye detection during night driving.

### 3.4.1. PCA based eye detection

Haar-like features fail to provide a reliable accuracy in the context of eye detection as the Haar-like features present in the eye are less prominent as compared to face. Hence, PCA [64] based eye detection has been adopted for day driving conditions owing to its superior real-time performance compared to other methods as reported in [65]. The ROI is resized to a resolution of 200×70 using bi-cubic interpolation [56]. Block wise search is carried out in each sub-window of size 50×40 pixels with a 10% overlap. The training algorithm is stated briefly in Table 3-10.

**Table 3-10: PCA based Eye Training Algorithm**

| |
|---|
| 1. Obtain eye images $I_1, I_2, I_3 \dots I_P$ (for training), each of dimension $N \times M$ |
| 2.  Represent every image $I_i$ of that class as a vector $\Gamma_i$ (of dimension $NM \times 1$) |
| 3.  Compute the average eye vector using $$\psi = \frac{1}{P} \sum_{i=1}^{P} \Gamma_i$$ |
| 4. Subtract the mean eye from each image vector $$\Gamma_i \phi_i = \Gamma_i - \psi$$ |
| 5. The estimated covariance matrix, $C$ is given by: |

$C = \frac{1}{P} \sum \phi_n \phi_n^T = AA^T$. Where $A = [\phi_1, \phi_2... \phi_P]$ ($N*M \times P$ matrix)

As C is very large, compute $A^T A$ ($P \times P$) instead as $P \ll N$

6. Compute the eigenvectors $v_i$ of $A^T A$.

$$\sigma_i u_i = A v_i$$

Using the equation above Eigenvectors $u_i$ of $AA^T$ are obtained

7. Depending on computational capacity available, keep only $K$ eigenvectors corresponding to the $K$ largest Eigen values. These $K$ eigenvectors are the Eigen eyes corresponding to the set of $M$ eye images

8. Normalize these $K$ eigenvectors

The value of K for face detection was selected as 15 empirically after observing the Eigen values. The training is carried out using 460 eye images cropped from some images taken from Dataset I each of size 50×40. The detection is carried out by the method as in Table 3-11.

**Table 3-11: PCA based Eye Detection Algorithm**

1. The ROI obtained from detected face, is resized to 200×70 matching with the resolution of training images

2. 10% overlapping windows, $\Gamma$ of size 50×40 are used for testing. Then, $\phi$ is obtained from the mean image $\psi$ as

$$\phi = \Gamma - \psi$$

3. Compute

$$\widehat{\phi} = \sum_1^k w_i \, u_i$$

4. The error, $e$ is computed as

$$e = \| \phi - \widehat{\phi_J} \|$$

5. The window corresponding to the minimum error, $e$ is selected as eye.

To examine the performance of eye detection, a test was conducted with 300 face images having 200 open and 100 closed eyes taken from Dataset I but different from those used in training. Fig. 3-11 shows some Eigen eyes obtained from the training set. Fig. 3-12 shows some detection results using PCA. From Table 3-12, it is clear that the PCA technique is quite accurate for detecting the eye from the ROI.

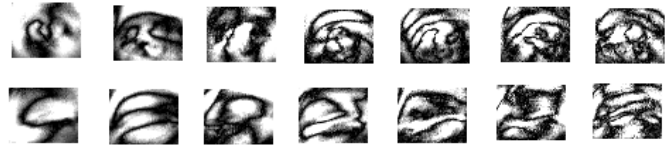

**Fig. 3-11 Sample Eigen eyes**

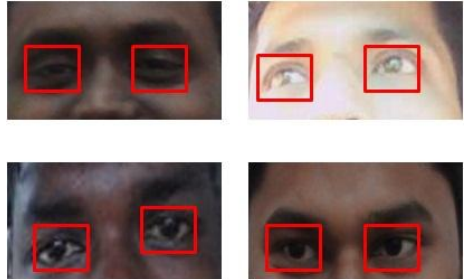

**Fig. 3-12 Sample Eye detections using PCA**

**Table 3-12: Eye detection results using PCA**

| tp | tn | fp | fn | tpr | fpr |
|----|----|----|----|-----|-----|
| 198 | 97 | 3 | 2 | 98.5% | 2.02% |

## 3.5  Detection of eye in NIR images

The PCA based method failed to be robust during night driving conditions, where passive NIR lighting is used to illuminate the face. A low cost Gallium Arsenide NIR illuminator is designed for the purpose. The Block-LBP histogram [66] features is used for detection of eyes in NIR images. Illumination invariance property of LBP features is a major advantage in using them for this purpose.

### 3.5.1. LBP features

LBP features are popularly used in object detection due to its discriminative power and computational simplicity [66]. The LBP operator labels the pixels of an image by thresholding the neighborhood of each pixel and considering the result as a binary number. The middle point in a 3×3 neighborhood is set as the threshold, $i_c$ and it is compared with the neighborhood intensity values, $i_n$ for $n = 0, ... ,7$ to obtain the binary LBP code.

$$LBP = \sum_{n=0}^{7} s(i_n - i_c) 2^n \qquad\qquad 3\text{-}16$$

Where, $s(x) = \begin{cases} 1, x \geq 0 \\ 0, x < 0 \end{cases}$. \qquad\qquad 3-17

### 3.5.2. Block LBP histogram

For eye detection, a global description of eye region is required. This is achieved using Block LBP histogram. In the LBP approach for texture classification, the occurrences of the LBP codes in an image are collected into a histogram. The classification is then performed by computing simple histogram similarities. But these result in loss of spatial information. One way to overcome this is the use of LBP texture descriptors to build several local descriptions of the eye and combine them into a global description. These local feature based methods are more robust against variations in pose or illumination than holistic methods.

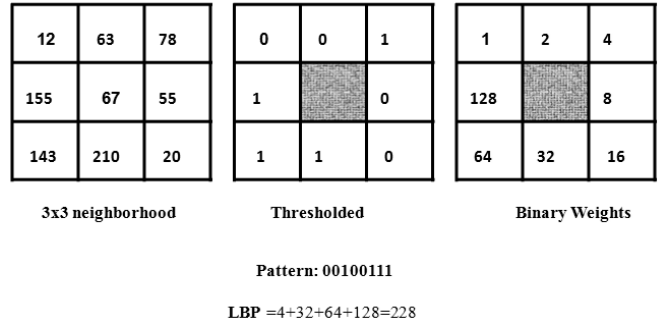

**3x3 neighborhood**       **Thresholded**       **Binary Weights**

**Pattern: 00100111**

**LBP** =4+32+64+128=228

**Fig. 3-13 LBP Calculation**

The algorithm for LBP based face description which was proposed by Ahonen *et al.* [66] is as follows:

- The image of the object (face) is divided into local regions
- LBP texture descriptors are extracted from each region independently
- The descriptors are concatenated to form a global description of the face

For eye detection, the training of LBP features has been carried out from eye images cropped out of the face images taken from Database III.

### 3.5.3. Eye Detection Algorithm using block LBP Features

The use of local histograms and pixel level values of LBP to get the feature vector is given in Table 3-13. In the detection phase, LBP histogram features are obtained in the localized eye region and then projected into the PCA feature space to get the energy components.

**Table 3-13: Algorithm: Block LBP**

| Training Phase |
|---|
|     1. Eye images are resized to 50×40 resolution and each image is divided into sub-blocks of size 5×4 |
|     2. The LBP feature values of the sub-blocks are found and 16 bin histogram of each block is calculated |
|     3. The histogram of each block is arranged to form a global feature descriptor |
|     4. PCA is carried out on the feature vectors and 40 Eigen vectors having largest Eigen values were selected. |
| Detection Phase |
|     1. ROI from face detection stage is obtained |
|     2. For each sub window the Block - LBP histogram is found and is projected to Eigen space |
| 3. The sub window with minimum reconstruction error is found |
| 4. If the reconstruction error is less than a threshold it is considered as a positive detection |

The algorithm is tested on the frontal upright as well as in-plane rotated face images extracted from Database III. The figure which follows shows some eye detection results using block LBP features.

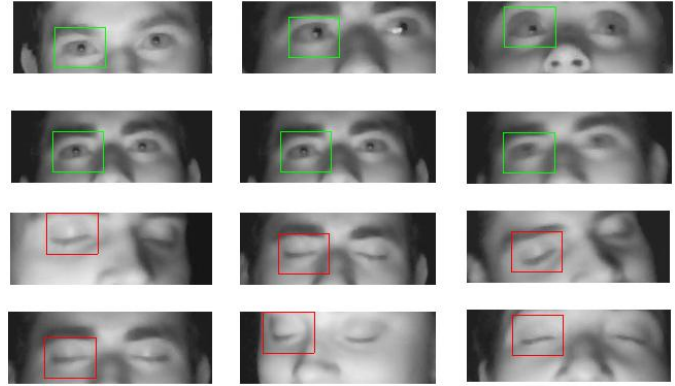

**Fig. 3-14 Eye localization using LBP**

The results as in Table 3-14 reveal its robustness over PCA for NIR images. However, as evident from Table 3-15, PCA performs faster than block LBP feature based method, and has been used for eye detection during day driving.

**Table 3-14: Run-time Comparison of PCA and Block LBP**

| Method | Speed (fps) |
|---|---|
| PCA | 7.8 |
| LBP | 6.6 |

**Table 3-15: Comparison of PCA and Block LBP for NIR images**

| | tpr | fpr |
|---|---|---|
| PCA | 80% | 10% |
| LBP | 98% | 2% |

## 3.6 Eye state classification

For accurate estimation of PERCLOS, the detected eye needs to be accurately classified as open or closed states. SVM is reported to be a robust classifier for a two-class problem [67]. In the present case, the eye state classification is a two-class classification problem where the class labels include open and closed eye classes and hence SVM is used. The training of SVM has been carried out with 460 images (230 open and 230 closed eyes) taken from the database created using normal (Database I) and NIR illumination (Database III). Some training images are shown in Fig. 3-15. The testing is carried out using another 1700 images taken from the same sources. Now the weight vectors, corresponding to each sample found in the eye detection phase, is fed to the SVM along with the ground truth. The training is carried out with different kernels and accuracy levels and the classification results are shown in Table 3-16.

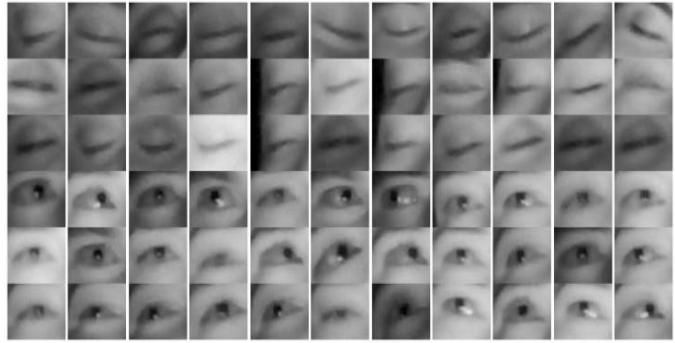

**Fig. 3-15 Training images for eye state classification using SVM**

**Table 3-16: Detection results with SVM**

| Kernel function | tp | tn | fp | fn | tpr | fpr |
|---|---|---|---|---|---|---|
| Linear SVM | 838 | 808 | 42 | 12 | 98.58% | 4.94% |
| Quadratic | 827 | 765 | 85 | 23 | 97.29% | 10% |
| Polynomial (order 3) | 848 | 807 | 43 | 2 | 99.76% | 5.32% |

The results reveal that the performance of the SVM is consistent under normal and NIR images. The number of support vectors is 16. A linear kernel will be better in terms of memory requirement. However, a third order polynomial kernel is found to provide the most accurate result and hence been implemented in the final algorithm.

Once the eyes are classified, the algorithm computes the PERCLOS value using the number of open and closed eye count over a sliding time window of three minutes duration. The real-time algorithm is deployed into an SBC having Intel Atom processor, with 1.6 GHz processing speed and 1 GB RAM. The overall processing speed is found to be 9.5 fps, which is quite appreciable for estimating PERCLOS.

### 3.7 PERCLOS estimation – A Case Study for Automotive Drivers

This section presents a case study for drowsiness detection in automotive drivers using PERCLOS. The above algorithm for estimating PERCLOS is tested under laboratory conditions and validated using EEG signals. However, on-board conditions are quite different from the laboratory conditions and therefore testing in practical driving scenarios is necessary. Hence, on-board tests were performed separately for day and night to test the robustness of the algorithm in actual driving conditions.

### 3.7.1. System Description

The alertness monitoring system consists of a single USB camera having maximum frame rate of 30 fps at a resolution of 640×480 pixels. The camera is placed directly on the steering column just behind the steering wheel to obtain the best view of the driver's face. The SBC is powered at 12 V DC from the car supply. The typical current drawn from the input source is approximately 1200 mA. The approximate typical power drawn from the supply is 15 W. A voltage regulator unit, comprising of IC LM317 along with some resistors, a capacitor and an inductor, is used before the input to remove high voltage spikes from the car supply. An NIR lighting arrangement, consisting of a matrix of 3×8 Gallium Arsenide LEDs, is also connected across the same supply in parallel with the Embedded Platform. The NIR module is operated at 10 V DC and draws a typical current of 250 mA. The lighting system is connected through a Light Dependent Resistor (LDR), to automatically switch on the NIR module in the absence of sufficient illumination. A seven inch LED touch screen is used to display the results. Fig. 3-16 shows the different modules used in the system.

**Table 3-17 Specifications of the Components**

| Component | Features | Typical power drawn |
|---|---|---|
| Embedded Processing Unit | 1.6 GHz, x86 | 9W |
| Voltage regulator unit | IC LM317 | |
| USB Camera | RGB, 640 x 480 @ 30fps | 1W |
| NIR lighting system | 3×8 Gallium Arsenide LEDs | 2W |
| LCD screen | 7", Resistive touch | 1W |
| Speaker | | 2W |

### 3.7.2. Testing under Day-driving Conditions

The main objective of the testing was to check the performance of the system at different lighting conditions during day time. The testing was carried out on a rough road followed by a smooth road (highway). The face and eye detection was found to be accurate for both smooth as well as jerky road conditions. But during the midday, when the light fell directly on the camera lens, the camera sensor got saturated and image features were lost. The problem was rectified later by using a camera with a superior image sensor to remove the saturation effect.

Fig. 3-18 shows a sample detection result during the start of the test. The analysis of the accuracy of the system has been carried out offline on set of 1000 images selected from the on-board recorded videos to evaluate the performance of the system. The face hit rate, when the driver was looking straight, is found to be 98.5% whereas the eye hit rate was 97.5% on an average. With head rotations of the driver, the face hit rate dropped down to 95% whereas that for eyes was 94%. The eye state classification was found to be 97% accurate. The overall false alarming rate of the algorithm was found to be around 5%.

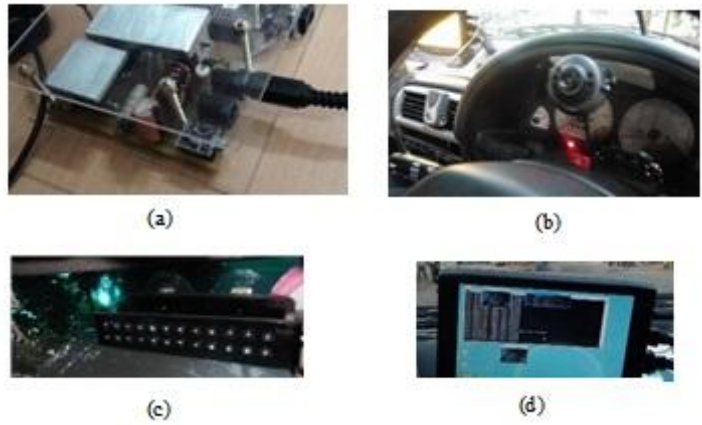

**Fig. 3-16 (a) surge protector circuit (b) camera on dashboard
(c) NIR lighting system (d) LCD screen on dashboard**

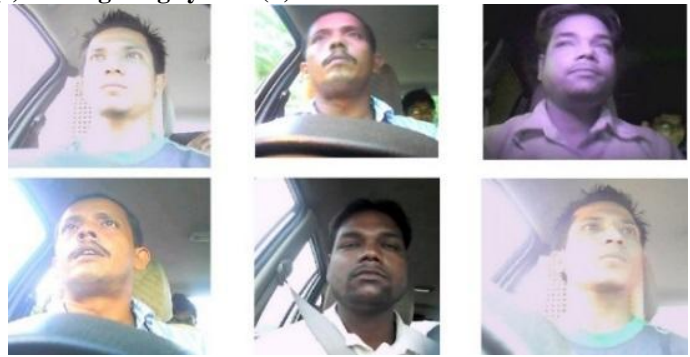

**Fig. 3-17 On Board testing with different subjects**

The error in PERCLOS estimation was calculated by the difference of true PERCLOS computed manually with the PERCLOS value obtained from the algorithm. The error is found out to be 4.5% approximately. The PERCLOS values are shown in the display for every minute, on a running average window basis of 3 minutes. The main issues faced during the day driving test are the image saturation due to sunlight directly falling on the camera lens and noise due to vehicle vibration. There is still scope of research left in addressing the issues.

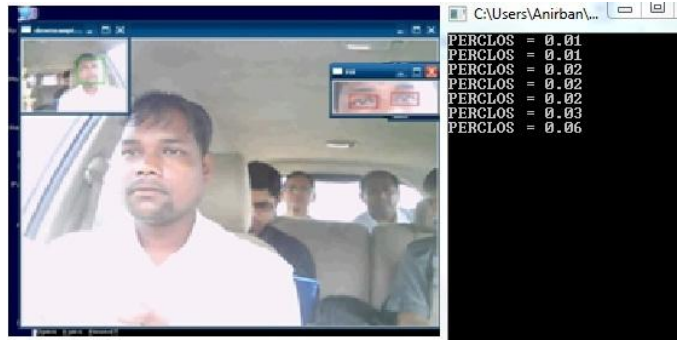

**Fig. 3-18 On-board operation of the system during daytime**

### 3.7.3. Testing under Night Driving Conditions

As per literature, most of the accidents due to drowsiness of the driver occur during the night [68] [69]. Hence, separate on-board testing during night was carried out. The NIR lighting system is automatically switched on by the LDR. The illumination level was enough to detect faces properly providing a clear NIR image without distracting the driver. The tests were performed on a highway. The objectives of the test was to check the performance of the NIR lighting system and to find out the performance of the algorithm in detecting face and eyes during night driving condition. Except for a few cases of malfunctioning owing to the lights coming from the other vehicles and the streetlights, the system was found to be robust in detecting the face and eyes. The face detection rate, when the driver was looking straight, is found to be 98.2% whereas the eye detection rate was 97.1%. With head rotations of the driver, the face detection rate dropped down to 94.5% whereas that for eyes was 94.2%. The eye state classification was found to be 98% accurate. The overall false alarming of the algorithm was found to be around 5.5%. The error in PERCLOS estimation was calculated by the difference of true PERCLOS computed manually with the PERCLOS value obtained from the algorithm. The error is found out to be 4.9% approximately. Fig. 26 shows a detection during night driving. The NIR lighting system works well for the test duration. From the above on-board tests, it is evident that this system can be used as a safety precaution system which might prevent a number of road accidents due to drowsiness. The results are summarized in the following table.

**Table 3-18: Summary of results**

| Day Driving | | |
|---|---|---|
| looking straight | Face hit rate | 98.5% |
| | Eye hit rate | 97.5% |
| head rotations of the driver | Face hit rate | 95% |
| | Eye hit rate | 94% |
| eye state classification | | 97% |
| overall false alarming rate | | 5% |
| error in PERCLOS estimation | | 4.5% |
| **Night Driving** | | |
| looking straight | Face hit rate | 98.2% |
| | Eye hit rate | 97.1% |
| head rotations of the driver | Face hit rate | 94.5% |
| | Eye hit rate | |

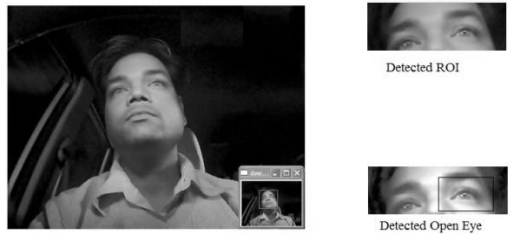

**Fig. 3-19 Detection of eyes in NIR lighting**

## 3.8 Conclusion

In this chapter, a real-time algorithm to estimate PERCLOS has been discussed. Face detection using Template Matching, PCA and Haar-like features has been compared and Haar based detection is preferred over Template Matching and PCA owing to its better accuracy and real-time performance. A real-time framework is presented where the input frame is down sampled for low resolution face detection, leading to extraction and remapping of ROI for eye detection. An optimal down sampling SF of 5 is selected as a trade-off between accuracy and run-time. The affine transformation of the input frame is carried out to detect in-plane rotated faces. Illumination compensation is carried out using BHE. The eye are localized in real-time using PCA based method during normal illumination and using LBP features under NIR illumination. The eye state has been classified as open or closed using SVM based classifier. A case study of drowsiness detection in automotive drivers has been presented based on PERCLOS estimation. A system has been developed and tested for real-time assessment of drowsiness in automotive drivers.

# Chapter 4. Correlation of Image and EOG based saccadic parameter estimation

In the previous chapter, it is seen how PERCLOS has been computed in real-time. Nevertheless, PERCLOS is a detector of loss of attention due to drowsiness and hence there is need for other measures for early detection of loss of attention. Saccadic eye movements are reported to be an earlier indicator of loss of attention as compared to PERCLOS [16]. Such eye movements are recorded using scleral search coils, EOG and high speed videos of eye image sequences [70]. Scleral search coil based and EOG based methods are accurate, but being contact-based method have limited feasibility of implementations in practical situations. Image-based method, being a non-contact method, has high feasibility of implementation. In this chapter, a correlation study has been presented for image and EOG based saccadic parameter estimation, so that non-contact image based method may replace contact-based accurate methods such as EOG.

## 4.1 Experiment Design

An experiment was conducted to simultaneously record EOG as well as video of eye images. The details of the video capture has been provided in Chapter 2 under Database IV. In this section, the EOG capture and processing is provided in brief.

### 4.1.1. EOG Capture

The bipolar EOG electrodes were placed on distal ends of the forehead, beside the corner of the eye. The ground and reference electrodes were placed on the middle of the forehead. The subjects were asked to place their chin on a fixed base such that movement of the face becomes limited to decrease the occurrences of artefacts. The EOG signal, which is basically the Corneal Retinal Potential (CRP), is acquired using a poly-somnograph signal acquisition system utilizing the one horizontal ($EOG_H$) and two vertical ($EOG_V$) bi-channels. The $EOG_H$ electrodes are placed on the distal ends of the forehead, by the corner of the eyes. This bi-channel electrode measures the horizontal eye movements, characterized by high peaks during pure directional horizontal saccades. The $EOG_V$ electrodes are placed on each eye where one terminal remains same as $EOG_H$, whereas the other terminal is placed below the eye. This bi-channel produces considerable peaks during eye lid movements such as, Eye closure and openings as well as blinks. Fig. 4-1 shows the EOG electrode placements in the face of a subject.

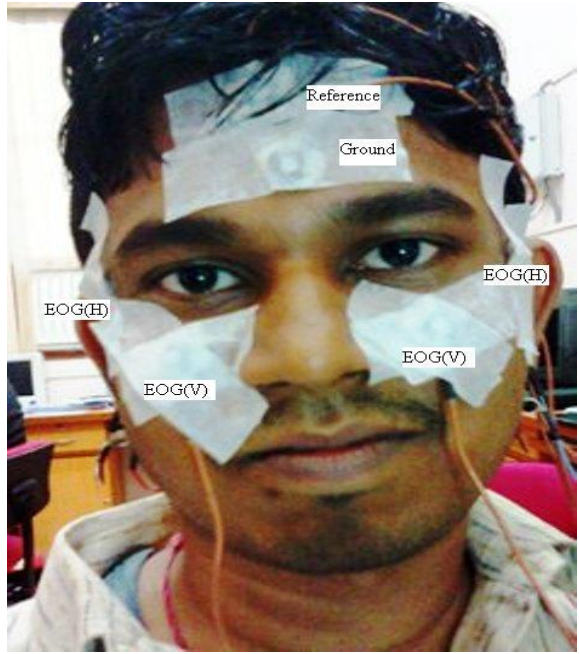

**Fig. 4-1 EOG Electrode Placements**

The EOG data in the bi-channel EOG$_H$ electrode were directly stored in computer through USB interface. The subjects were seated in a height-adjustable chair to keep their faces aligned with camera. The sampling frequency of the signal acquisition system hardware was 256 Hz.

### 4.1.2. EOG signal processing

The raw EOG data contained noise in the high frequency band (70-100 Hz). To remove this noise, a band-pass filter was used to limit the signal in 0.4 to 30 Hz band.

The filtered data was then normalized by removing the mean as given in 4-1.

$$d(n) = r(n) - \frac{1}{N}\sum_{i=1}^{N} r(i) \qquad \text{4-1}$$

where $d(n)$ is the normalized data, $r(n)$ is the raw data, $N$ is the length of the raw series.

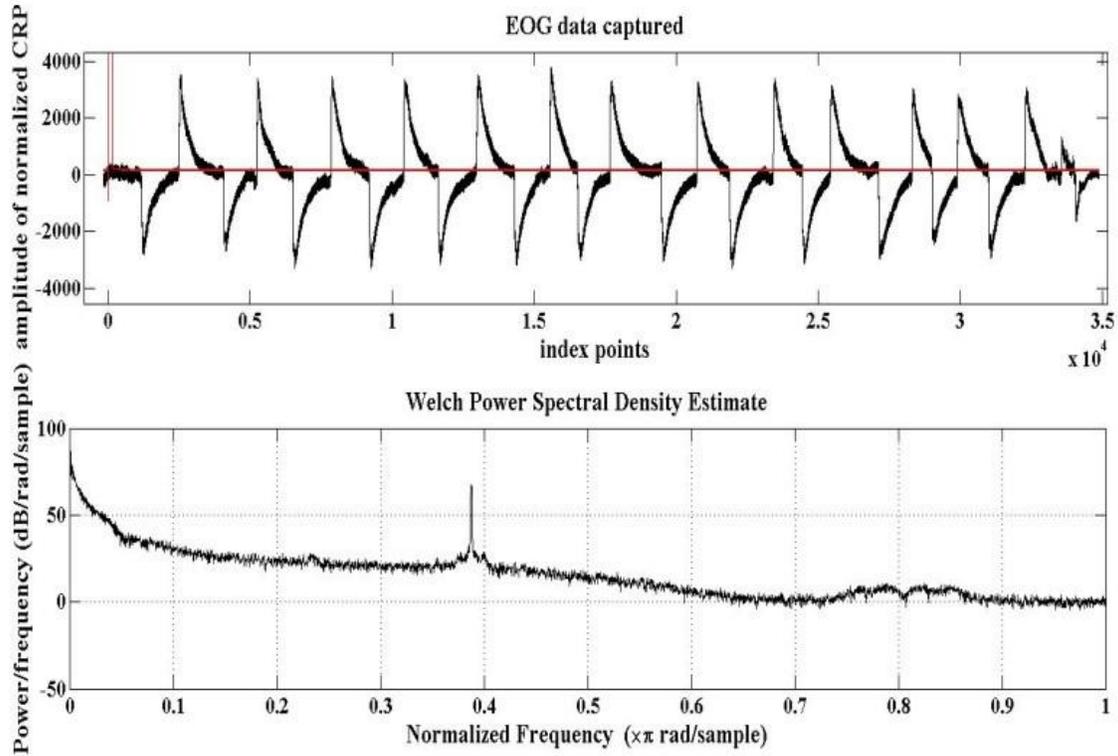

**Fig. 4-2 Raw EOG Data and its PSD**

### 4.1.3. Truncation of the data

==The index points in the figure indicate the sampled EOG data points. The data from horizontal EOG channel is shown in Fig. 4-2.== It could be clearly observed in Fig. 4-2 that the CRP displays clear peaks during pure horizontal eye saccades positive in case of right to left movement and negative in case of the opposite movement. ==The peaks were isolated by selecting a hard threshold (15% of peak CRP) and replaced these instantaneous data values with 0, using 4-2. The value 15% was obtained empirically.== Here, $m(n)$ is the truncated sequence.

$$m(n) = \begin{cases} d(n) & when \ |d(n)| > d_{thr} \\ 0 & when \ |d(n)| < d_{thr} \end{cases} \qquad 4\text{-}2$$

### 4.1.4. Expression of the Data in per unit term

For the purpose of comparing with the video data, the EOG data is represented in per unit terms $s(n)$.

$$s(n) = \frac{m(n)}{|\max(m(n))|} \qquad 4\text{-}3$$

### 4.1.5. Isolation and Parameterization of the Saccadic Peaks

The peaks were then identified by recognizing the gaps. The negative saccadic peaks were also converted to absolute values, with appropriate flags attached.

Fig. 4-3 shows the truncated EOG data along with its Power Spectral Density (PSD). The following parameters were identified from the peaks –

- The maximum CRP obtained during a Peak (Saccadic Amplitude)= $s(n)$
- The Difference series of the CRP series (Saccadic Velocity) $v(n)$= $s(n) - s(n-1)$
- The Maximum Value of Difference in CRP (Peak Saccadic Velocity)=max[$v(n)$]

## 4.2 Parameter comparison

The parameters mentioned above were calculated separately from video and EOG data. The pupil center was manually marked for each frame along with the temporal eye corner. The difference between the positions provided the saccadic amplitudes. The raw EOG data is a time-series whose value gives the potential at the horizontal electrode w.r.t. the reference electrode. The value of this potential is an indicator of horizontal eye position/movement. The saccadic velocities were obtained using the difference is saccadic amplitudes scaled by the frame rate. For comparison of the data, each parameter obtained from video data was also expressed in per unit terms.

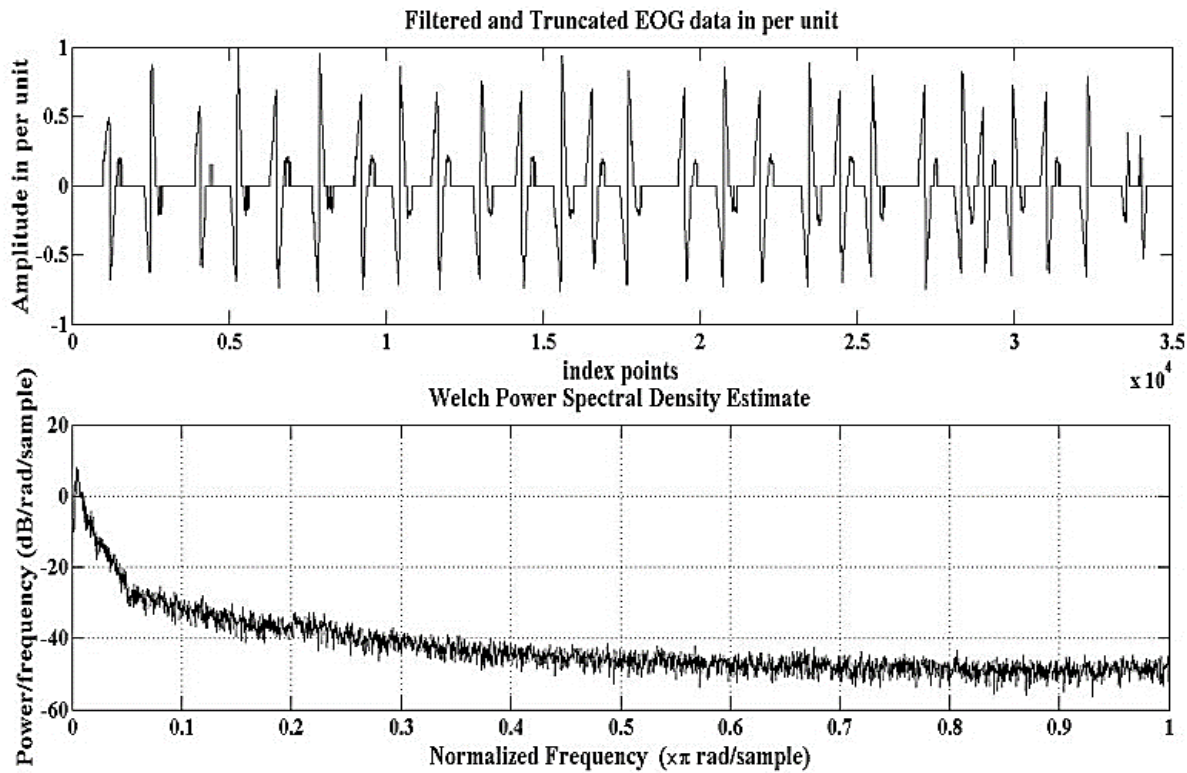

Fig. **4-4** shows a saccadic amplitude variation as obtained from the EOG data. Fig. 4-5 gives the pupil position data with respect to the eye corner along with its PSD.

**Fig. 4-3 Truncated EOG Data and its PSD**

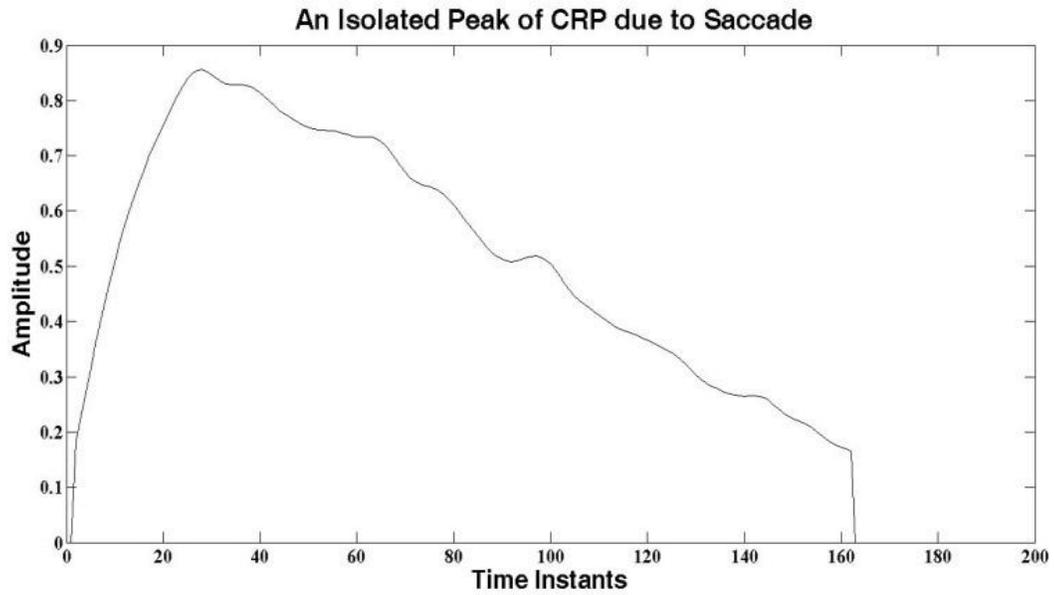

**Fig. 4-4 Isolated Saccadic Peak obtained using EOG data**

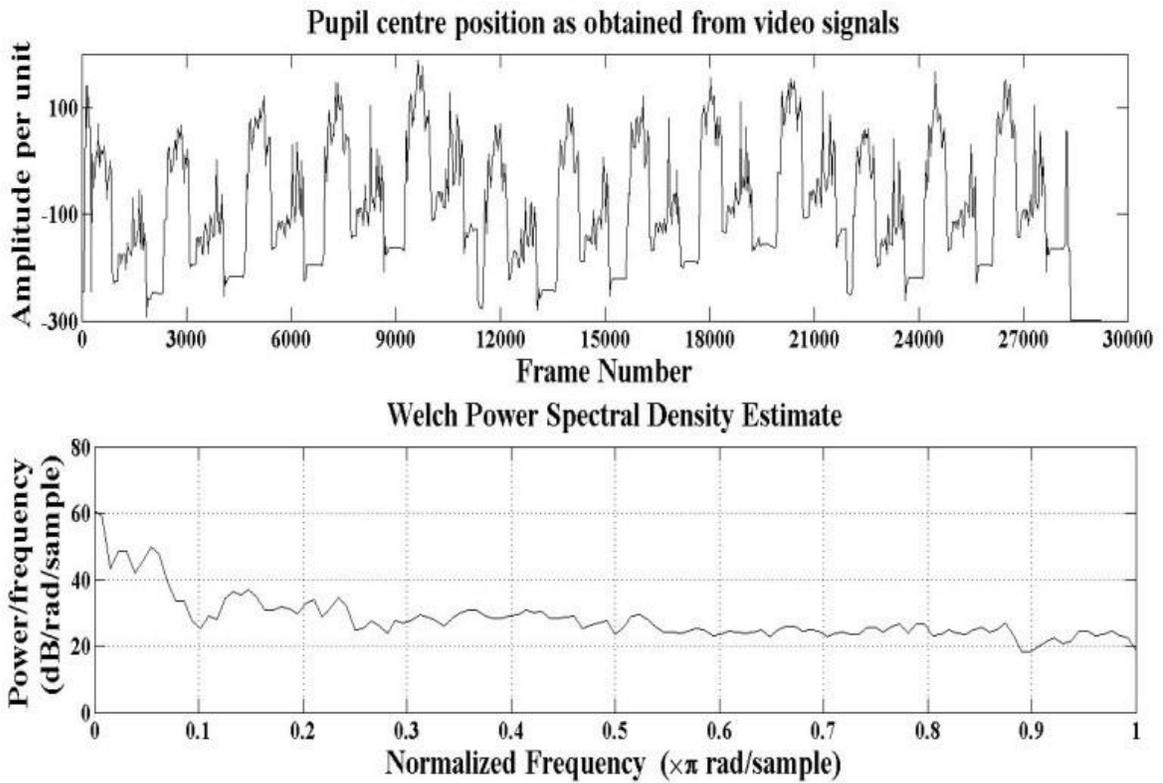

**Fig. 4-5 Pupil center position as obtained from video signals**

**4.3 Results**

Fig. **4-6** shows the comparison of peak saccadic amplitude as obtained using video data as
well as the EOG data. Fig. 4-7 shows the same for the peak saccadic velocity. The graphs
show a lot of similarity with some randomness in situations introduced due to noise.

**Fig. 4-6** shows the Amplitude of the SR's obtained from the EOG and Video. The correlation
of these 4 features has been found, using Pearson's linear correlation coefficient, to be 0.4066
of Peak Saccadic Amplitude, whereas the Peak Saccadic Velocity from the two methods had a
correlation of 0.7701.

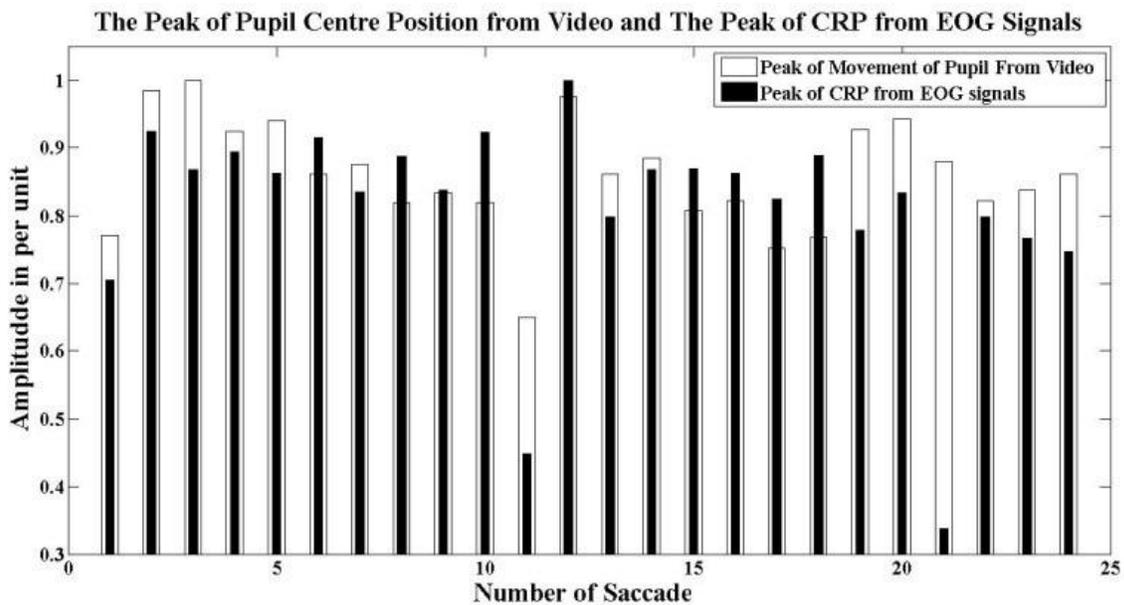

**Fig. 4-6 Comparison of peak saccadic amplitude by EOG and video**

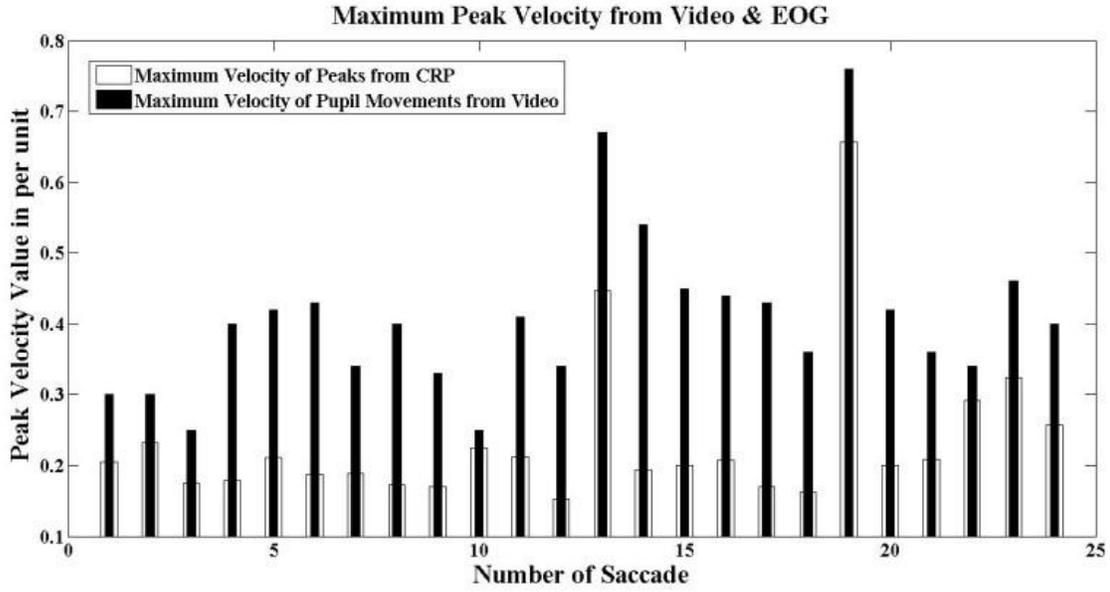

**Fig. 4-7 Comparison of peak saccadic velocity by EOG and video**

The work has been published in [71].

## 4.4 Discussion

The main objective of this work is to correlate the image based saccadic parameter estimation with that of EOG based, so that the non-contact image based method may replace the contact-based EOG based method. The peak amplitude and peak velocities of saccades have been plotted as observed by EOG as well as video data to compare and correlate the two acquisition methods. The pattern of the saccadic parameters as obtained by EOG as well as video is similar except at few instances because of noise in the data. The results convey that the contact based EOG technique may be replaced by the non-contact image based assessment. The main challenge remains in image based assessment is the fast and accurate detection of iris/pupil center.

# Chapter 5. Saccadic Ratio Estimation

From the previous chapter it is apparent, that image based estimation of saccadic parameters is correlated with that of more accurate EOG based estimation. This chapter aims at image based Saccadic Ratio (SR) estimation. SR may be defined as below

$$SR = \frac{V_p}{T_s} \qquad \text{5-1}$$

In the above equation, $V_p$ is peak saccadic velocity and $T_s$ is saccadic duration.

The challenges in image based saccadic parameter estimations are:

- Removal of glint from eye image for accurate iris detection
- Lack of standard video databases of eye saccades at high frame rates
- Lack of fast algorithms for accurately detecting the iris
- Processing at sufficiently high frame rates
- Noisy images are captured at high frame rate due to low exposure

The saccadic parameter estimation may be carried out accurately at sufficient frame rates by modeling the iris motion and tracking the pupil center. Several tracking methods exists in literature such as mean shift tracking [72], optical tracking [73] and Kalman Filter (KF) based tracking [74]. The first two methods rely on image intensity values in subsequent frames and hence are dependent on illumination levels thereby making them unsuitable for the application.

In the problem of iris tracking, the iris center is confined within the eye corners. Hence, in this work, a KF with constrained states has been implemented to estimate the SR.

## 5.1 Saccadic Parameters

Literature [2] reports that the peak saccadic velocity to duration ratio i.e. SR is an indicator of alertness level. For the estimation of SR, some saccadic parameters are needed to be pre-estimated which are defined as follows:

*Saccadic Amplitude* - This is the angular displacement of the iris with respect to the eye corner, when the subject executes a saccade. Saccadic amplitude $\theta$ is computed as the difference between the pupil centre position, $\theta_p$ and the eye corner position, $\theta_c$ as given below.

$$\theta = \theta_p - \theta_c \qquad \text{5-2}$$

In this work, the amplitude $\theta$ is measured as a fraction of the total eye width, to account for the distance from the camera i.e. pixels to angle mapping.

*Peak Saccadic Velocity* - This is the maximum velocity reached during a saccade. Saccade velocity profiles are usually symmetrical for small and medium saccades [75]. Peak saccadic velocity $\omega_p(k)$ at $k^{\text{th}}$ frame is computed as the maximum of the difference in pupil positions in consecutive frames over the saccadic duration, scaled by the frame rate, $fps$ as shown below.

$$\omega_p(k) = \max[(\theta(k) - \theta(k-1)) * fps] \qquad 5\text{-}3$$

*Saccadic Duration* - This is the time taken to complete the saccade. This is most easily measured from the velocity profile. Saccadic duration, $t_s$ is computed as

$$t_s = \frac{k_{\max[\theta]} - k_{min[\theta]}}{fps} \qquad 5\text{-}4$$

Here, $k_{\max[\theta]}$ denotes the frame where $\theta$ reaches maximum while $k_{min[\theta]}$ denotes the frame where $\theta$ reaches minimum.

## 5.2  Iris Detection

For estimating the saccadic parameters, it is important to detect the iris center accurately. The eye is localized using methods as described in Chapter 3. The detected eye region is gamma corrected as proposed in [76] to compensate the effects of illumination variations, local shadowing, and highlights as well as to preserve the original visual appearance. The local dynamic range of the eye image gets enhanced in dark regions while compressing it in bright regions. In this work, use of Projection Functions [77] and Form Factor [3] have been tested for detecting the iris position. Thereafter, a novel framework is proposed for iris detection.

### 5.2.1. Projection Functions

Image projection functions have been used in [77] for localizing the eye region. From the results of the work in [77] Generalized Projection Function (GPF) has been used for locating the iris center in the eye image $I(x, y)$. $IPF_v$ and $IPF_h$ are the vertical and horizontal integral projection functions respectively. $VPF_v$ and $VPF_h$ are the vertical and horizontal Variance Projection Functions respectively. $GPF_v$ and $GPF_h$ are the vertical and horizontal GPF's respectively.

$$IPF_v(x) = \frac{1}{y_2 - y_1} \sum_{y_1}^{y_2} I(x, y) \qquad 5\text{-}5$$

$$IPF_h(y) = \frac{1}{x_2-x_1}\sum_{x_1}^{x_2} I(x,y) \qquad\qquad 5\text{-}6$$

Tha variance projection function, $VPF_v$ are computed as

$$VPF_v(x) = \frac{1}{y_2-y_1}\sum_{y_1}^{y_2}[I(x,y) - IPF_v(x)] \qquad\qquad 5\text{-}7$$

$$VPF_h(y) = \frac{1}{x_2-x_1}\sum_{x_1}^{x_2}[I(x,y) - IPF_h(y)] \qquad\qquad 5\text{-}8$$

Finally, the vertical and horizontal GPF's are obtained as

$$GPF_v(x) = (1-\alpha)IPF_v(x) + \alpha VPF_v(x) \qquad\qquad 5\text{-}9$$

$$GPF_h(y) = (1-\alpha)IPF_h(y) + \alpha VPF_h(y) \qquad\qquad 5\text{-}10$$

As the method in [77], with a setting of $\alpha = 0.6$, the iris centre is obtained. Some iris centre detections are shown below.

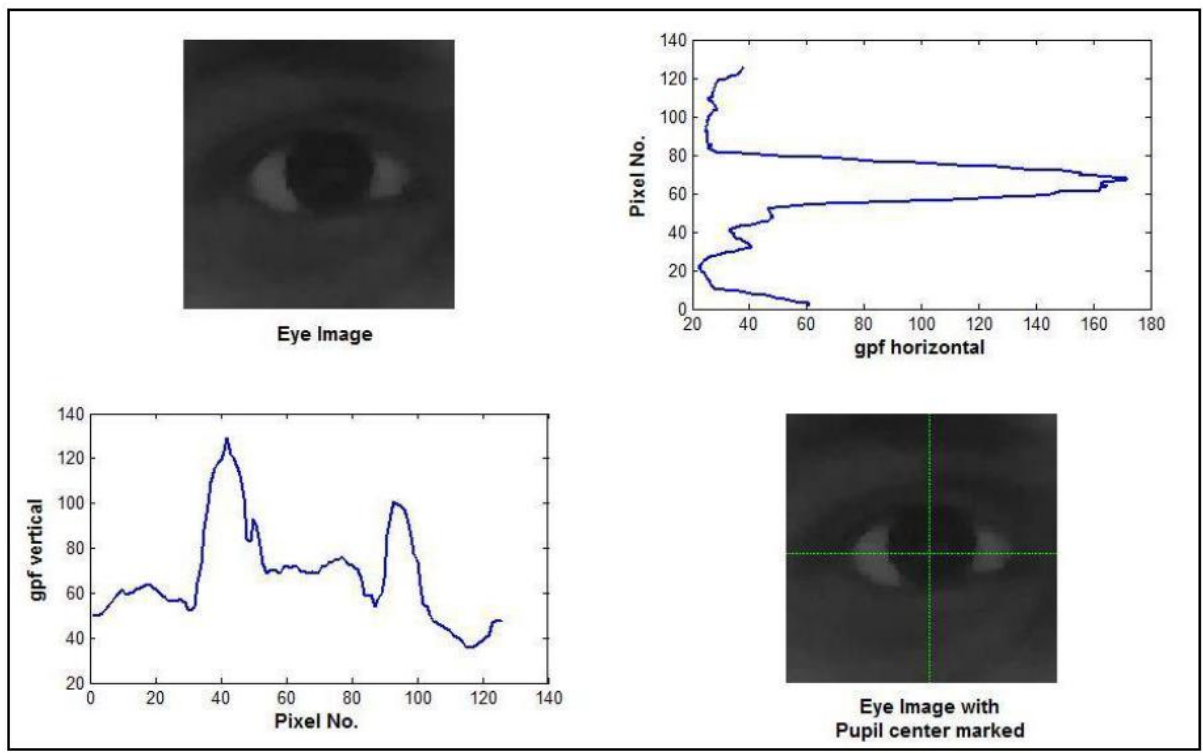

**Fig. 5-1 Iris detection using GPF**

In the above figure, it can be observed that there are two large peaks in $GPF_v$ which correspond to the two iris edges. The mid point of the location of the peaks gives us the x-coordinate of pupil center. The $GPF_h$ has a single large peak, which gives us the y-coordinate of pupil center. The estimated pupil center has been shown on to the original image.

The projection functions for pupil detection was computationally inexpensive. A frame rate of 20 fps was obtained on eye video databases (Database IV). The method is most accurate when the iris is located at the eye centre. The algorithm was found to be inaccurate in case of variation in illumination as well as eye gaze.

## 5.3 Form Factor based detection

Form Factor (FF) based iris detection was achieved by Gupta *et al.* [78]. The algorithm detects the possible position of the iris by computing the FF at every pixel of the image. The FF at a pixel is defined as the ratio between Root Mean Square (RMS) to average value of pixel intensities in the region around it. In this work, a rectangular neighbourhood region of 3×3 pixels centred at the pixel of interest has been considered to be its surrounding pixels. The expression for FF at a point is given by:

$$\mathbf{FF} = \frac{R.M.S}{Average} = \sqrt{\left(1 + \frac{\sigma^2}{\mu^2}\right)} = \sqrt{N}\frac{\|x\|_2}{\|x\|_1} \qquad 5\text{-}11$$

where, $R.M.S$ is the RMS value of pixel intensities in the neighbourhood, $Average$ is the average pixel intensity of the neighbourhood, $\mu$ and $\sigma$ are the mean and standard deviation of pixel values in the region respectively. $\|.\|_p$ is the $p$ norm of the pixel sequence of $N$ pixels around it. This ratio provides two measures of the image. One is the contrast and the other is the volume of low intensity pixels. The FF defined in 5-11 is not a normalized quantity. Hence the inverse of square of FF is selected as a normalized index to quantify the image feature such as an edge [78]. Thus the Edge Strength Index (ESI) denoted by $\beta$ may be expressed as:

$$\boldsymbol{\beta} = \frac{1}{FF^2} = \frac{\|x\|_1^2}{\|x\|_2^2} \qquad 5\text{-}12$$

A higher value of ESI indicates an edge owing to the higher entropy in the region [78]. The value of the ESI calculated was scaled to a value from 0 to 255 and was used to create an edge map. By using Circular Hough Transform (CHT) [41] to detect circles, the possible iris centre is located.

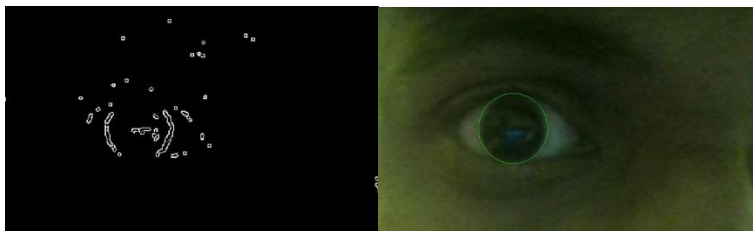

**Fig. 5-2 Iris detection using FF based method**

In the above figure, image on the left is the edge image generated using the ESI values and that on the right shows the detected iris. The above example shows a successful iris detection. However, there may be spurious low intensity pixels due to improper illumination which may result in incorrect determination of iris centre. The example below shows such a case.

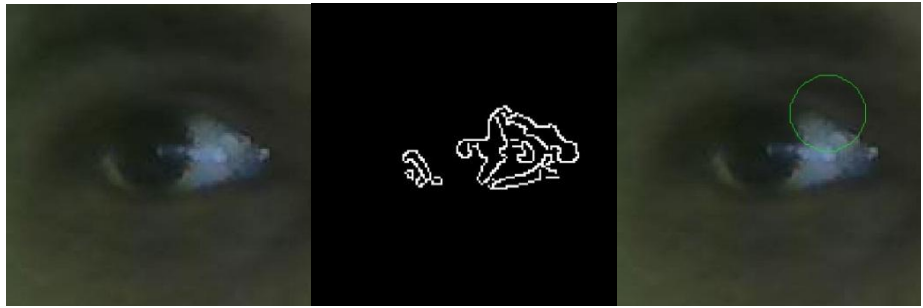

**Fig. 5-3 False detection using FF based method**

In the above figure, the left image is the input eye image, middle one being the edge image generated using the ESI values, while the right one shows the iris detection. It can be observed that all the desired edges are not present and thus this results in unsuccessful detection of iris. The performance of FF based approach is found to be dependent upon the illumination level as well as the eye gaze.

## 5.4 Novel Framework for Iris Detection

The issues observed with the above methods are inaccuracies in iris detection mainly caused due to lighting conditions and partial occlusion. A major issue is the glint on the image of the eye surface i.e. the reflection of ambient light on the pupil which creates a white patch on the black iris region. By performing morphological opening [56], the glint was found to be removed in most of the cases. The morphological opening involves morphological erosion followed by dilation, using the same structuring element for both operations as given below:

$$A \circ B = (A \ominus B) \oplus B \qquad\qquad 5\text{-}13$$

where $\ominus$ and $\oplus$ denote erosion and dilation, respectively. In the present work, the structuring element employed is a disk of radius 10 pixels.

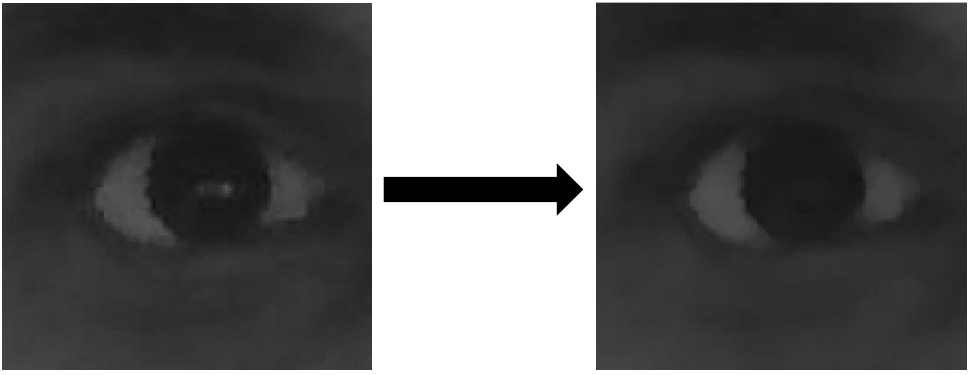

**Fig. 5-4 Removal of glint using morphological opening**

Then, the process of locating the iris centre involves extraction of the edge map, thresholding followed by iris detection using CHT. Due to the shape of the eye sockets and varying illumination, the Canny edge detection algorithm [56] is found to detect a lot of redundant edges in addition to the useful edges. The redundant edges are removed by using thresholding to obtain a reduced edge map.

The edge detection is carried out using four successive stages:

- Noise reduction
- Intensity gradient computation
- Non-maximum suppression
- Useful Edge Extraction

*Noise reduction:* The morphologically opened eye image is convolved with a Gaussian filter to reduce the high frequency noise. The filter matrix $B$ is given below:

$$B = \frac{1}{159}\begin{bmatrix} 2 & 4 & 5 & 4 & 2 \\ 4 & 9 & 12 & 9 & 4 \\ 5 & 12 & 15 & 12 & 5 \\ 4 & 9 & 12 & 9 & 4 \\ 2 & 4 & 5 & 4 & 2 \end{bmatrix} \qquad 5\text{-}14$$

*Finding the Intensity gradient:* The Canny algorithm uses four filters to detect horizontal, vertical and diagonal edges in the filtered image. The first derivative in the horizontal direction, $G_x$ and the vertical direction, $G_y$ are obtained by convolving the image with kernels $H_x$ and $H_y$ as shown below.

$$H_x = \begin{bmatrix} 1 & 0 & -1 \\ 2 & 0 & -2 \\ 1 & 0 & -1 \end{bmatrix} \qquad 5\text{-}15$$

$$H_y = \begin{bmatrix} 1 & 2 & 1 \\ 0 & 0 & 0 \\ -1 & -2 & -1 \end{bmatrix} \qquad 5\text{-}16$$

Using $G_x$ and $G_y$, the edge gradient, $G$ and direction, $\theta$ can be determined as given below:

$$G = \sqrt{G_x{}^2 + G_y{}^2} \qquad \qquad 5\text{-}17$$

$$\theta = \arctan\left(\frac{G_y}{G_x}\right) \qquad \qquad 5\text{-}18$$

The edge direction angle is rounded to one of four angles i.e. $0^0, 45^0, 90^0, 135^0$ representing vertical, horizontal and the two diagonals respectively.

*Non-maximum suppression*: After obtaining the edge gradient $G$ and direction $\theta$ a search is then carried out to determine if $G$ assumes a local maximum in the gradient direction $\theta$. The following rule is used:

**Table 5-1: Rule for non-maximum suppression**

| Rounded gradient angle | Gradient magnitude is greater than the magnitudes of pixels in |
|---|---|
| 0 degrees | north and south directions |
| 90 degrees | west and east directions |
| 135 degrees | north east and south west directions |
| 45 degrees | north west and south east directions |

*Useful Edge Extraction:* Some of the unnecessary edges have been removed by thresholding. Thresholding has been carried out using two thresholds high and low. The thresholds are obtained experimentally. This results in a reduced edge map.

On this reduced edge map, CHT with a predefined radius range is used to identify the iris location. The centre of the detected circle is selected to be the iris centre.

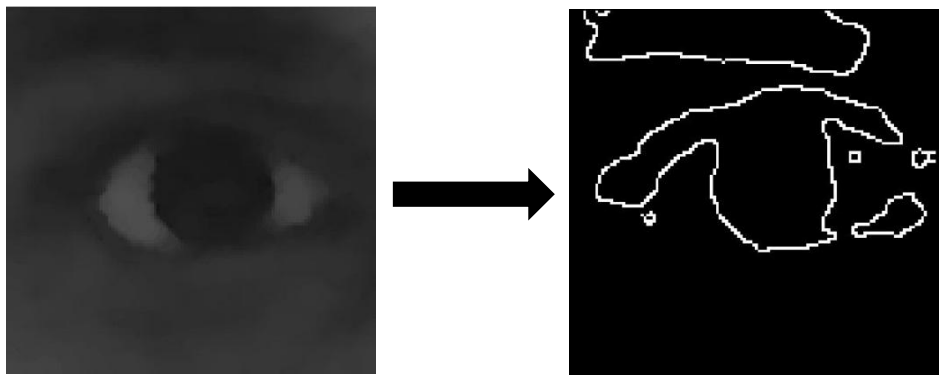

**Fig. 5-5 Edges detected from the morphologically opened eye image**

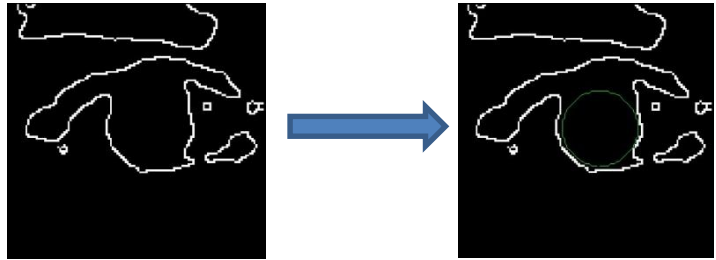

**Fig. 5-6 Iris detection using CHT on the reduced edge map**

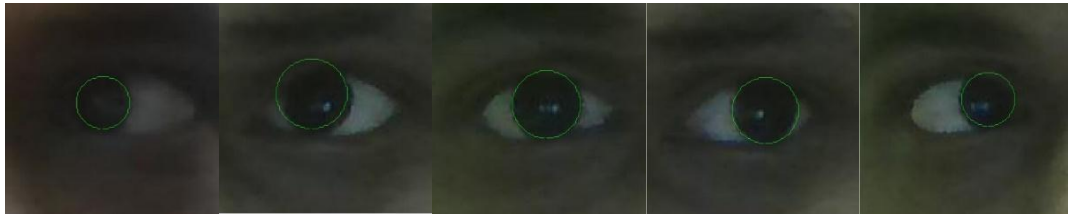

**Fig. 5-7 Iris detection with different eye gazes**

## 5.5 Eye corner detection

The eye corners have been used to serve as a reference point for computing the relative iris centre position. The eye corners have been selected as reference positions because the position of eye corners does not vary with different facial expressions, levels of eye closure, gaze, eyelashes or makeup. A corner is defined by the intersection of at least two edges. The work in [79] describes about detection of both the eye corners (temporal as well as nasal). In this work, we have used only the temporal eye corner of the left eye as the reference point. However, both the eye corners have been detected to obtain the saccadic amplitude as a fraction of the eye width, to account for scale-invariant position values.

## 5.6 KF based Tracking

The run-time performance of the algorithm decreases owing to the computational burden of the algorithm. This has been improved by using the temporal and spatial information. The presence of eye in the region is verified once every 6 frames and if this detection fails, face detection is carried out followed by eye detection on the whole image. As the iris centre is assumed to be known accurately in a frame, then the subsequent positions can be tracked using a KF. The KF also serves the purpose of filtering out the measurement and process noise. The iris centre measurements are not always perfect and the accuracy varies with lighting conditions. In certain frames, the measurement fails due to closure of eyelids or poor illumination levels. These erroneous measurements may be rectified by using the predicted states and using prior knowledge about the maximum possible eye movement.

In tracking, the pupil position is selected around the predicted position in the subsequent frame thereby reducing the tracking failure. The iris centre position obtained as stated above is used as the state vector along with the iris velocity, which is obtained from the difference of iris centre position in two consecutive frames.

### 5.6.1. System Model

Let $x_k$, where $k$ is the frame index, denote the true four-dimensional position and velocity state vector that is to be estimated. The model used for tracking is a discrete linear time-invariant system model given by:

$$x_{k+1} = Ax_k + Bu_k + w_k \qquad \text{5-19}$$
$$z_k = Hx_k + v_k \qquad \text{5-20}$$

Here, $u_k$ is the known control input, $z_k$ is the measurement. The process noise $w_k$ and measurement noise $v_k$ are assumed to be Additive White Gaussian Noise (AWGN) with zero mean and covariance given as

$$E\{w_k w_k^T\} = Q \qquad \text{5-21}$$
$$E\{v_k v_k^T\} = R \qquad \text{5-22}$$

$E\{\}$ is the expectation operator, $Q$ and $R$ are the covariance matrices for process and measurement noise respectively. The state transition matrix $A$, input transition matrix $B$ and output transition matrix $H$ are given by:

$$A = \begin{bmatrix} 1 & 0 & 1 & 0 \\ 0 & 1 & 0 & 1 \\ 0 & 0 & 1 & 0 \\ 0 & 0 & 0 & 1 \end{bmatrix} \quad B = \begin{bmatrix} T & 0 \\ 0 & T \\ T & 0 \\ 0 & T \end{bmatrix} \quad H = I \quad T = \frac{1}{frame\ rate} \qquad \text{5-23}$$

The Kalman estimate $\widehat{x_k}$ of $\widehat{x_k}$ is found using the following equations.

$$K_k = A\Sigma_k H^T (H\Sigma_k H^T + R)^{-1} \qquad \text{5-24}$$
$$\widehat{x_{k+1}} = A\,\widehat{x_k} + B\,\widehat{u_k} + K_k(Z - Hx_k) \qquad \text{5-25}$$
$$\Sigma_{k+1} = (A\Sigma_k - K_k Z\Sigma_k)A^T + Q \qquad \text{5-26}$$

$\hat{x}$ is the constrained Kalman estimate and $\Sigma$ is the covariance matrix of the estimate.

The original measurements are obtained by manually marking the pupil centre in a video of eye images recorded at 60 fps using a USB camera.

## 5.7 Constrained KF based Tracking

For linear dynamic systems with white process and measurement noises, the KF is known to be an optimal estimator [80]. In the application of KF there is often available model or signal information that is ignored. In a constrained KF, the filter is modified such that known inequality constraints are satisfied by the state-variable estimates.

Standard quadratic programming results are used to solve the KF problem with inequality constraints. At each time step of the constrained KF, a quadratic programming problem is solved to obtain the constrained state estimate. A family of constrained state estimates is obtained, where the weighting matrix of the quadratic programming problem determines which family member forms the desired solution. Of all possible constrained solutions has the smallest error covariance. In many practical problems, the states are constrained within some bounds as given:

$$d_{min,k} \leq Dx_k \leq d_{max,k} \qquad \text{5-27}$$

$D$ is an $s \times n$ constant matrix with $s$ no. of constraints and $n$ state variables with $s \leq n$. Considering $D$ is full-rank, the constrained KF problem can be solved by directly projecting the unconstrained state estimate $\tilde{x}$ onto the constraint surface. The constrained estimate is obtained by solving:

$$\min_{\tilde{x}}(\tilde{x}^T \Sigma^{-1} \tilde{x} - 2\hat{x}^T \Sigma^{-1} \tilde{x}) \qquad \text{5-28}$$

such that $d_{min,k} \leq Dx_k \leq d_{max,k}$ is satisfied. $\hat{x}$ is the constrained Kalman estimate and $\Sigma$ is the covariance matrix of $x$. The constrained state estimate $\tilde{x}$ is unbiased and has a smaller error covariance than the unconstrained estimate $\hat{x}$. The constrained Kalman filter shows better tracking performance when the predicted state exceeds the constraints due to erroneous measurements. The constrained Kalman estimate settles quicker than a normal Kalman estimate.

From the figure which follows, it can be observed that when the variable being measured goes beyond the constraints and comes back, the constrained KF settles much quicker than the unconstrained KF. At frame no. 63, the measurement comes within the left bound and constrained estimate settles by frame no. 66 whereas the unconstrained estimate settles at frame no. 72. Similarly, at frame no. 115, the measurement comes within the right bound and constrained estimate settles by frame no. 118 whereas the unconstrained estimate settles at frame no. 125.

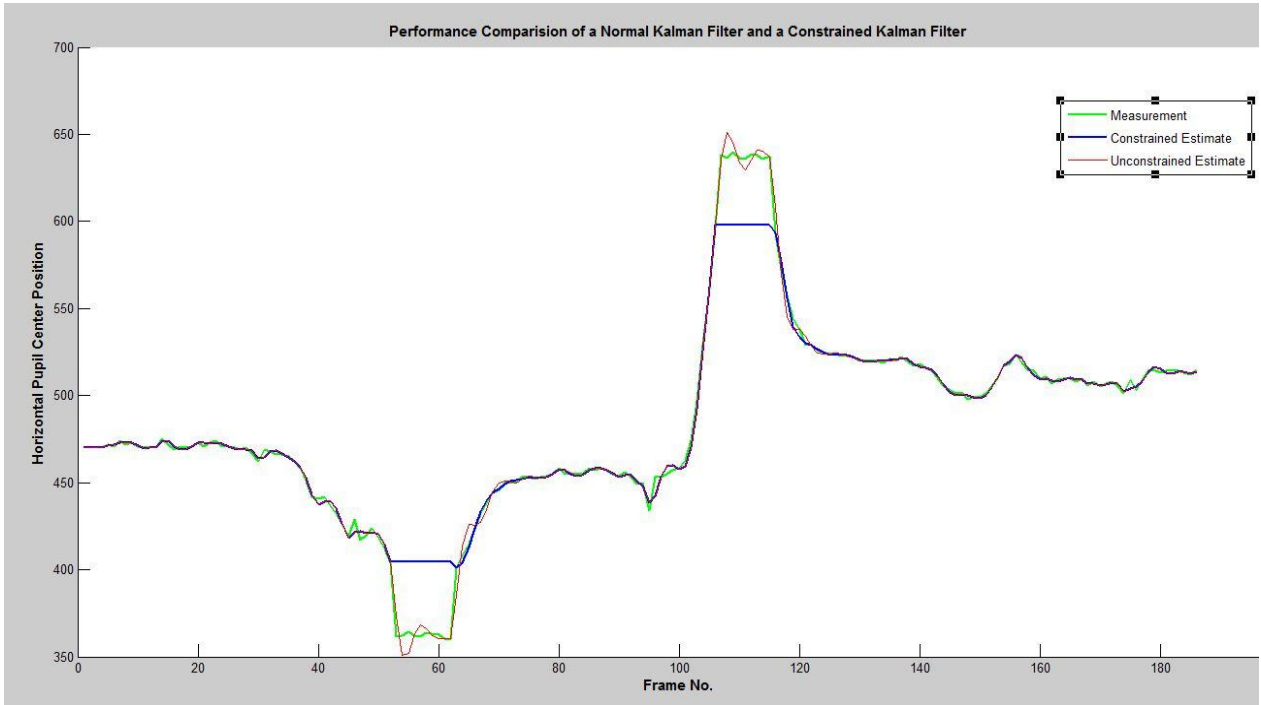

**Fig. 5-8 Performance comparison of Constrained and Unconstrained KFs**

The eye corner positions were used to determine the constraints. The superiority of constrained KF helps in achieving better tracking performance compared to an unconstrained KF especially when the measurement goes beyond the constraints and comes back into the normal range.

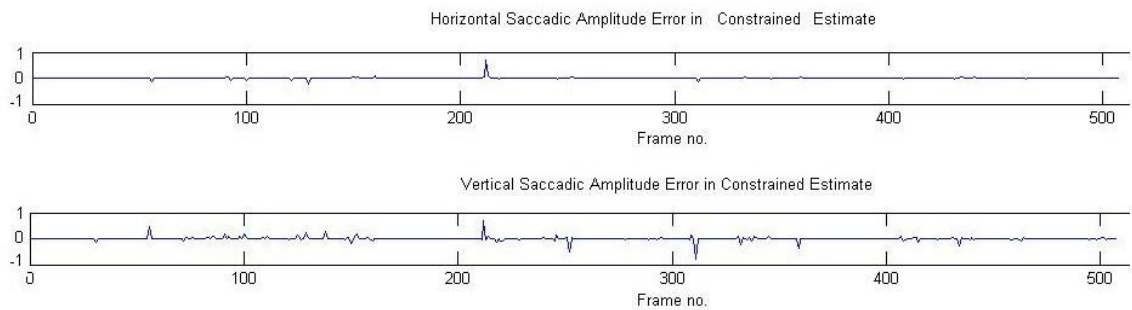

**Fig. 5-9 Error in Saccadic Amplitude Estimation**

## 5.8 SR Estimation

The SR has been estimated using the peak saccadic velocity and saccadic duration.

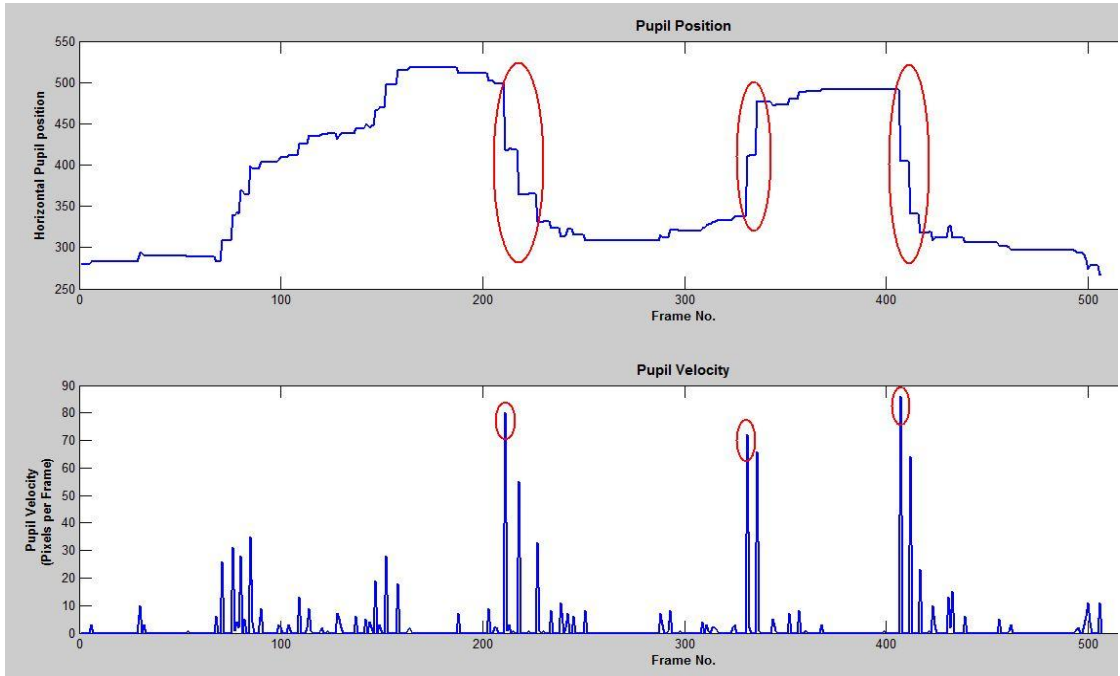

**Fig. 5-10 Duration and Peak Velocity Estimation**

The above figure shows the pupil position and velocity. The red ellipses indicate the saccades. Starting from left, the first saccade occurred between frames 210 and 227 with a peak velocity of 80 pixels per frame (ppf) and amplitude of 167 pixels (p). Second saccade occurred between frames 330 and 336 with peak velocity 72 ppf and amplitude 140p. Third saccade occurred between frames 406 and 417 with peak velocity 86 ppf and duration 173p. The above figure shows the pupil position and velocity for the test case. The red ellipses indicate the saccades. Starting from left, the first saccade occurred between frames 210 and 227 with a peak velocity of 80 ppf and amplitude of 167 pixels. Second saccade occurred between frames 330 and 336 with peak velocity 72 ppf and amplitude 140p. Third saccade occurred between frames 406 and 417 with peak velocity 86 ppf and duration 173p. The following Table gives the SR estimates for the test case and compares it with the ground truth for first 5 saccades. The overall error in SR estimation the algorithm as compared with manual marking of frames is found out to be approximately 12% as obtained. The algorithm executes at a speed of 30 fps on an Intel i5 processor with clock speed of 2.53 GHz. There is still scope of fine tuning of the algorithm by removal of noise due to low exposure time and high sampling rates.

**Table 5-2: Error in SR estimation**

| Saccade no. | SR estimate (pixels/ms$^{-2}$) | True SR (pixels/ms$^{-2}$) | % Error (True SR – estimated SR) in SR estimate |
|---|---|---|---|
| 1 | 0.821 | 0.771 | -6.48 |
| 2 | 0.561 | 0.682 | -17.74 |

| 3 | 0.702 | 0.651 | -5.1 |
|---|-------|-------|-------|
| 4 | 0.996 | 0.888 | -12.16 |
| 5 | 0.961 | 0.804 | -19.52 |

**5.9 Conclusion**

This chapter presents a framework for image-based estimation of SR. The issues prevailing in the literature are highlighted and some of them has been addressed. Two popular methods for detecting the iris are tested and a modified framework is proposed for detecting the iris. The eye corners are detected to obtain the iris center position as a fraction of the total eye width, to remove the effect of scaling. The temporal eye corner is selected as the reference point for estimating iris center position. As the iris center position being confined between the eye corners, a constrained KF has been deployed to track the iris center position. The SR is estimated from the iris position data. The algorithm executes at a speed of 30 fps on an Intel i5 processor with clock speed of 2.53 GHz and gives an error of around 12% for SR estimation. The speed of the algorithm may be increased by fixed point as well as parallel implementation on processors such as Graphics Processing Units (GPUs).

# Chapter 6. Detection of Spectacles

A major limitation in the eye detection methods stated in the earlier part of the thesis is the detection of eyes occluded by spectacles. The main issue in detecting the eyes for faces containing spectacles is the glint caused in the glasses, which results in loss of information regarding the state of eye. This chapter aims at design of an algorithm for detection of the presence of spectacles.

Several algorithms have been reported to detect the presence of spectacles in face images. In [81], Jiang *et al.* made an attempt to detect the presence of glasses using grey level discontinuities of the frame against the facial parts, with limited accuracy. In [82], Saito *et al.* attempted to detect eye glasses using PCA thereby removing them using Active Contour Model [83]. Similar attempts [84] , [85], [86] have been made to detect and remove spectacles, however, with confined accuracies.

In this chapter, a real-time algorithm has been proposed to detect the presence of spectacles, leading to the localization and detection of eyes.

## 6.1 The Algorithm

The algorithm may be divided into the following steps

1. Face detection
2. ROI selection
3. Contrast enhancement
4. Image Sharpening
5. Thresholding
6. Dilation
7. Detection of spectacles
8. Eye region extraction

### 6.1.1. Face detection and ROI selection

The face detection and ROI selection has been carried out using the method described in Chapter 3.

### 6.1.2. Contrast enhancement

Contrast enhancement is required for accurate detection of spectacles thereby making the detection illumination invariant. Adaptive Histogram Equalization (AHE) [87] is reported to be an efficient technique for improving the local contrast of an image. However, AHE has a property of over-amplifying noise in comparatively homogeneous regions of an image. Contrast Limited Adaptive Histogram Equalization (CLAHE) [88] removes this tendency by limiting the noise amplification. Hence, CLAHE has been applied to the ROI image as obtained in the previous stage.

### 6.1.3. Image Sharpening

Once a high contrast image has been obtained, the edges of the images are to be detected to determine the presence of spectacles. There are different masks [56] which are used for edge detection. There are two basic principles for each edge detector mask as given below:

1. The element values in the mask should sum to zero.

2. The mask should approximate differentiation or amplify the slope of the edge.

In this work, a 5×5 Laplacian operator has been used for image sharpening in the image which is given as

$$L = \begin{bmatrix} -1 & -1 & -1 & -1 & -1 \\ -1 & -1 & -1 & -1 & -1 \\ -1 & -1 & 24 & -1 & -1 \\ -1 & -1 & -1 & -1 & -1 \\ -1 & -1 & -1 & -1 & -1 \end{bmatrix} \qquad 6\text{-}1$$

This mask is convolved with non-overlapping windows of size 5×5 in the ROI image to obtain the sharpened image.

### 6.1.4. Thresholding

The edge-detected image is converted into a binary image using Otsu's Thresholding technique [89]. Otsu's method is applied to automatically execute histogram shape-based image thresholding. The algorithm presumes that the image to be thresholded contains two classes of pixels i.e. bi-modal histogram and subsequently computes the optimum threshold separating those two classes so that their combined spread (intra-class variance) is minimal. Finally, the edge detected region is thresholded as given below.

$$g(x, y) = \begin{cases} 1 \ if \ f(x, y) \geq T \\ 0 \ if \ f(x, y) < T \end{cases} \qquad 6\text{-}2$$

Here, $f(x, y)$ is a pixel intensity value of the image which is thresholded with a threshold $T$ to obtain the final binary image $g(x, y)$.

### 6.1.5. Dilation

All the connected components which have a size of less than a threshold are removed and the rest of the connected components are retained. Dilation [56] is carried out to connect any possible linkages using a small structuring element.

### 6.1.6. Spectacle Detection

The spectacle detection is performed based on a threshold called Detection Factor $D$, which is calculated using the longest connected component $l$ and the breadth $b$ of the face image as given:

$$D = \frac{l}{1.5 \times b} \qquad 6\text{-}3$$

Eventually, the spectacles are said to be detected based on $D$ according to the rule given below:

$$Output = \begin{cases} Detected, if \ if \ D \geq 1 \\ Not \ detected \ if \ D < 1 \end{cases} \qquad 6\text{-}4$$

### 6.1.7. Eye Region Extraction

Once the spectacles have been detected, the eye region is extracted using Distance Transform [90]. For each pixel in the binary image, the distance transform assigns a number that is the distance between that pixel and the nearest nonzero pixel in the binary image. This entails that each pixel value contains its distance from the non-zero pixel. As in the present case, the location of spectacles is already obtained as non-zero pixels, distance transform forms a suitable approach for this particular problem. The problem of eye region extraction is solved as the pixels having distance less than a threshold value fall into the eye region. The threshold value, $d_m$ is obtained as in 6-5 based on experimental results.

$$d_m = 0.8 * l \qquad \qquad \text{6-5}$$

The various types of distances [90] which have been calculated between pixels $(x_1, y_1)$ and $(x_2, y_2)$ to examine the best eye region extraction method are given in Table 6-1. Fig. 6-1 shows the comparison of results with different distance transforms.

**Table 6-1: Types of Distances**

| | |
|---|---|
| Chess-Board | $max\ (|x_1 - x_2|, |y_1 - y_2|)$ |
| City Block | $|x_1 - x_2| + |y_1 - y_2|$ |
| Euclidean | $\sqrt{(x_1 - x_2)^2 + (y_1 - y_2)^2}$ |
| Quasi-Euclidean | $\begin{cases} |x_1 - x_2| + (\sqrt{2} - 1)|y_1 - y_2|, if\ |x_1 - x_2| > |y_1 - y_2| \\ (\sqrt{2} - 1)|x_1 - x_2| + |y_1 - y_2|, if\ |x_1 - x_2| \le |y_1 - y_2| \end{cases}$ |

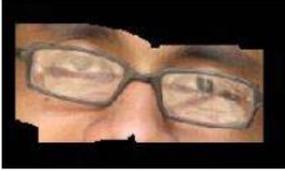

Chess board

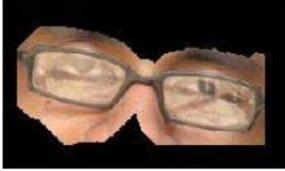

City Block

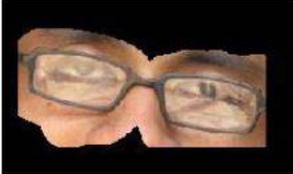

Euclidean

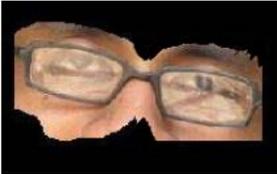

Quasi Euclidean

**Fig. 6-1 Results with different Distance Transforms**

## 6.2 Results

The algorithm is tested on Database V. Fig. 6-2 shows the detection of spectacles for a subject wearing spectacles. Different images are shown corresponding to each step as discussed in the algorithm. Fig. 6-3 reveals that the algorithm concludes presence of no spectacles in the face for a detection factor of less than unity. The overall performance of the algorithm is examined by plotting the ROC curve as shown in Fig. 6-4. The area under the curve is 0.9582 which reveals that the algorithm is robust for the database prepared. The algorithm executes a speed of 7 fps on an embedded platform having 1 GB RAM, 1.6 GHz clock speed, RT Linux.

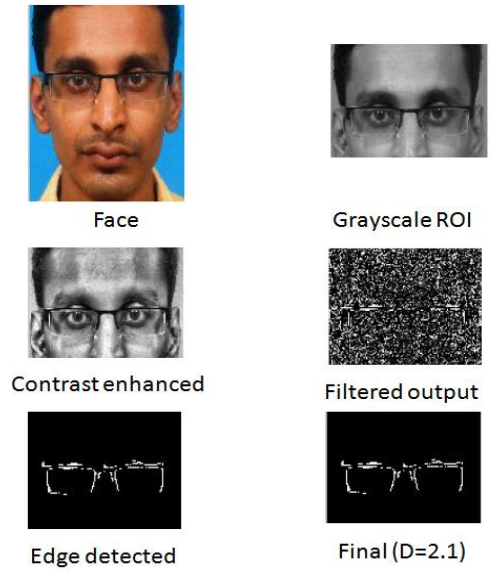

**Fig. 6-2 Different stages of spectacles detection**

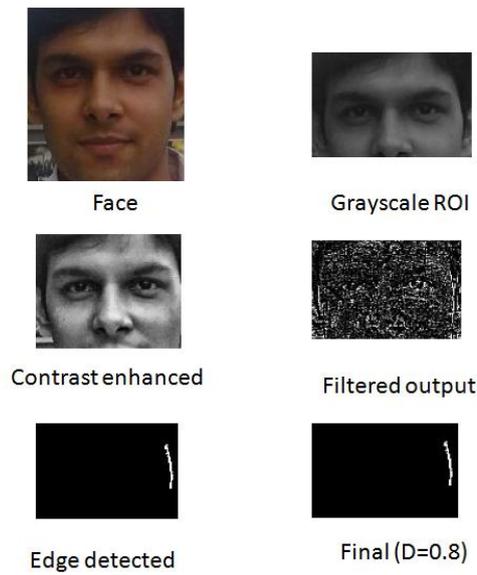

**Fig. 6-3 No spectacles detected with *D*<1**

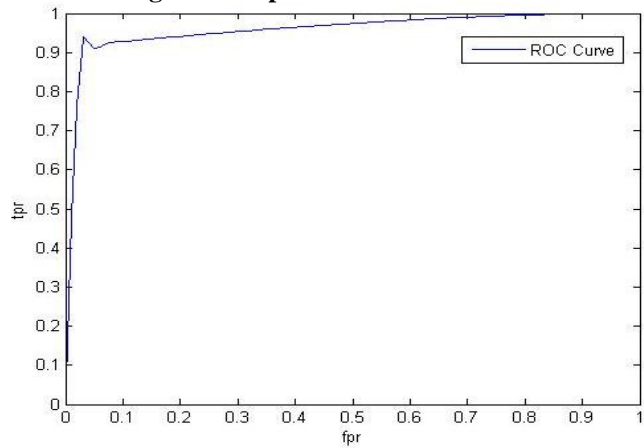

**Fig. 6-4 ROC curve for the algorithm on the database**

## 6.3 Major Issues during Spectacle Detection with Solutions

The major issues which have been identified during the evaluation of the performance of the algorithm are as follows:

### 6.3.1. Localized Standard Deviation

Standard deviation of the $3 \times 3$ neighborhood around the corresponding pixel in the input image gives the localized results. The output image was subsequently normalized by finding out the maximum value of the output image and dividing the output by the same. Fig. 6-5 shows the comparison of output of the operations using a $3 \times 3$ neighborhood around a pixel to find the standard deviation of that pixel. The variation in histogram exhibits the effect of using the standard deviation based approach.

### 6.3.2. Laplacian Operator

The Laplacian operator used for edge extraction has not been found to be robust in extracting all the features properly for the entire test database. Hence the operator had to be fine-tuned.

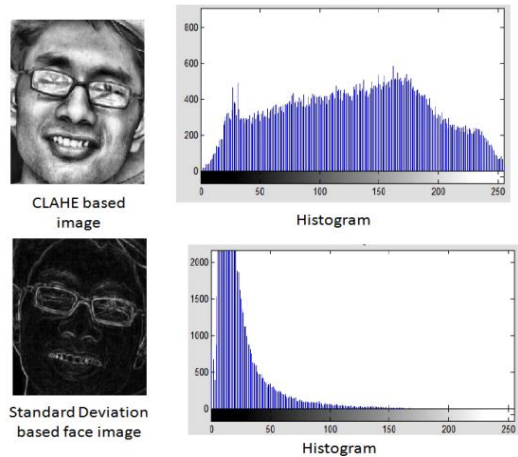

CLAHE based
image

Histogram

Standard Deviation
based face image

Histogram

**Fig. 6-5 Effect of using Standard Deviation prior to filtering**

As an image enhancement algorithm, the Difference of Gaussians (DoG) [91] has been utilized to increase the visibility of edges. Edge-sharpening filters which operate by enhancing high frequency detail, cause undesirable artifacts owing to the fact that random noise also has a high spatial frequency. The DoG algorithm removes high frequency detail that often includes random noise, rendering this approach one of the most suitable for processing images with a high degree of noise.

$$M = \begin{bmatrix} 0 & 0 & -1 & -1 & -1 & 0 & 0 \\ 0 & -2 & -3 & -3 & -3 & -2 & 0 \\ -1 & -3 & 5 & 5 & 5 & -3 & -1 \\ -1 & -3 & 5 & 16 & 5 & -3 & -1 \\ -1 & -3 & 5 & 5 & 5 & -3 & -1 \\ 0 & -2 & -3 & -3 & -3 & -2 & 0 \\ 0 & 0 & -1 & -1 & -1 & 0 & 0 \end{bmatrix} \qquad 6\text{-}6$$

The mask shown above is called "DoG" or "Mexican hat" having a positive peak at the centre (i.e. 16 in the 7×7 mask). Fig. 6-6 shows that the DoG operator brings out more features as compared to Laplacian operator.

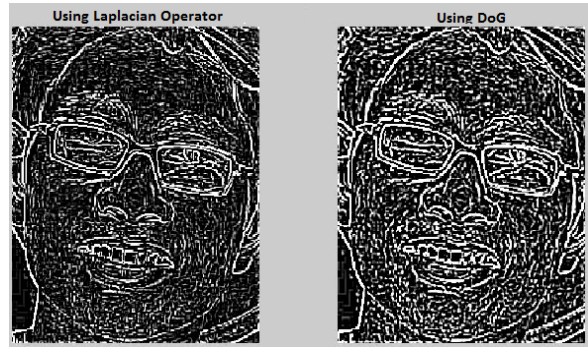

**Fig. 6-6 Comparison of Laplacian and DoG based edge detection**

### 6.3.3. Removal of Smaller Connected Components

All the smaller connected components were removed and only the four highest connected components were retained. This was carried out as the extracted features were still not connected even after dilation in certain cases. Increasing the size of structuring element, beyond a certain size, leads to inclusion of more noise. With proper experimentation using different connected components, the best four connected components corresponding to four lines in the spectacles (two upper and two lower have been selected.

The detection factor has also been modified, taking into account the best two connected components. If $l_1$ and $l_2$ are the size of the two best connected components and $b$ is the breadth of the face image, then the detection factor is given as

$$D = \frac{l_1 + l_2}{1.5 \times b}$$
6-7

This modification also minimizes the false alarm rate of spectacle detection due to some erroneous feature extraction. The work has been published in [92]

## 6.4 Conclusion and Future Work

This chapter presents a real-time algorithm for detection of spectacles leading to the detection of eyes. In this work, the face has been detected using Haar-like features and an ROI has been selected from the detected face region. The edges in the ROI image have been detected using a Laplacian mask. The edge detected image has been thresholded and subsequently dilated to detect the presence of spectacles. For eye region extraction Quasi-Euclidean distance transform has been found to be a better method than Chess Board, City Blocks and Euclidean. A database of video has been created for evaluating the performance of the algorithm as well as for aiding future researchers for detection of eyes in faces occluded with spectacles. The problems in the algorithms have been discussed and fine tuning of the algorithm has been carried out to improve the accuracy. There is further scope of research in the improvement of accuracy of spectacle detection. Removal of glint caused by the glasses is another potential future scope of this research.

# Chapter 7. Conception of the overall system

The conception of the combined PERCLOS and SR based alertness monitoring system is provided in this chapter. The system will record the facial images at high frame rates and detect the eyes and iris position to estimate the SR. In the event, the SR falls below a threshold, the system will provide an indication of reduced alertness level and start computing the PERCLOS value based on the eye closure rate. In case the PERCLOS is within threshold, the system will resume estimating the SR, otherwise when the PERCLOS reaches the threshold, the system will alarm the user as of being drowsy.

The figure below shows the schematic of the concept system with combined PERCLOS and SR as indicators of alertness levels. The main issue in the realization of such a system is hardware limitation which may be feasible in the future.

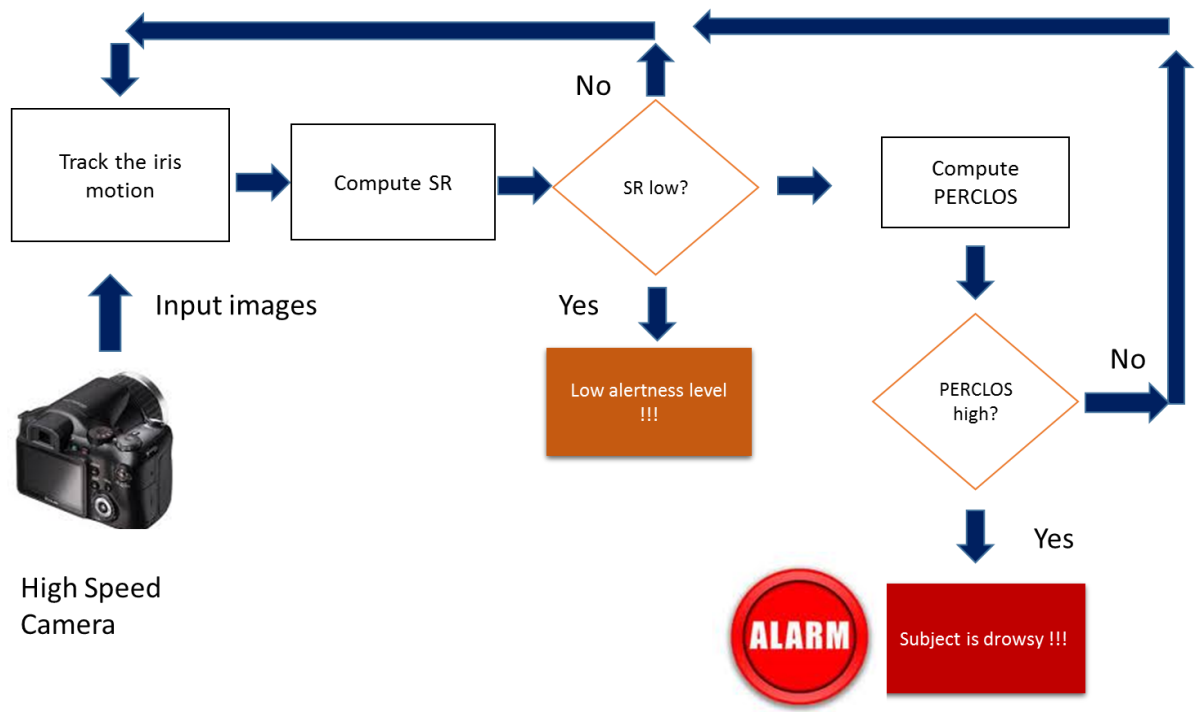

**Fig. 7-1 Conception of the overall System**

## 7.1 Conclusion

The thesis is about development of fast algorithms for estimating two ocular parameters – PERCLOS and Saccadic Ratio using techniques of computer vision. These parameters have been reported to be highly correlated with human alertness level. Hence, the purpose of fast computation of these parameters is to develop a product which will be used for gauging the level of alertness. Such a device is intended to prevent many accidents and hazards in certain safety and critical operations. PERCLOS has been tested in real-time and cross-validated with EEG signals. A case study on real-time monitoring of automotive drivers has been presented in which an embedded system based on PERCLOS has been designed. A correlation study of image and EOG based saccadic parameter estimation has been carried out to replace the contact based method using non-contact image based method.

Saccadic parameter estimation using image based methods has been addressed and a semi-real-time implementation has been discussed. An initial attempt has been made to detect the presence of spectacles in the face image, leading to eye detection.

Five sets of video databases have been created for evaluation of performances of the algorithms as well as for the research community for aiding future research.

## 7.2 Future Scope

The speed of the algorithm for estimating the saccadic parameters may be improved by a parallel implementation in platforms such as Graphics Processing Units (GPUs). There may be improvement in the algorithm for detection of eyes wearing spectacles. Moreover, glint removal is one more issue which needs to be addressed adequately.

Dynamic models of eye saccadic parameters may be formulated to study the evolution of loss of attention of an individual with respect to saccadic parameters. For real-time saccadic parameters estimation, advanced signal processing tools may be useful such as compressed sensing.

# Publications from the Thesis

**Journal Papers**

1. **A. Dasgupta**, A. George, S. L. Happy, A. Routray, "A Vision Based System for Monitoring the Loss of Attention in Automotive Drivers", in *IEEE Transactions on Intelligent Transportation Systems*, vol. 14, no. 4, pp.1825-1838, 2013
2. **A. Dasgupta**, A. George, S. L. Happy, A. Routray, Tara Shanker "An on-board vision based system for drowsiness detection in automotive drivers" in *International Journal of Advances in Engineering Sciences and Applied Mathematics, Springer, Special Issue on Advanced Traffic and Transportation Systems*, vol. 5, no. 2-3, pp. 94-103, 2013

**Patents**

1. A. Routray, **A. Dasgupta**, A. George, S. L. Happy, "A System for Real Time Assessment of Level of Alertness" (Patent Filed: 634/KOL/2013)

**Papers Presented at Conference Proceedings**

1. S. Gupta, **A. Dasgupta**, and A. Routray, "Analysis of Training of Parameters in Haar-like Features to Detect Human Faces," in *IEEE International Conference on Image Information Processing*, Shimla, India, 2011.
2. M. Singhvi, **A. Dasgupta**, A. Routray, "A Real Time Algorithm for Detection of Spectacles Leading to Eye Detection," *IEEE Proceedings of 4th International Conference on Intelligent Human Computer Interaction*, Kharagpur, India, 2012
3. A. Chaudhuri, **A. Dasgupta**, A. Routray, "Video and EOG Based Investigation of Pure Saccades in Human Subjects," *IEEE Proceedings of 4th International Conference on Intelligent Human Computer Interaction*, Kharagpur, India, 2012
4. S. L. Happy, **A. Dasgupta**, A. George, A. Routray, "A Video Database of Human Faces under Near Infra-Red Illumination for Human Computer Interaction Applications," *IEEE Proceedings of 4th International Conference on Intelligent Human Computer Interaction*, Kharagpur, India, 2012